\documentclass[11pt]{article}

\usepackage[preprint]{acl}
\usepackage{times}
\usepackage[utf8]{inputenc} % allow utf-8 input
\usepackage[T1]{fontenc}    % use 8-bit T1 fonts
\usepackage{url}            % simple URL typesetting
\usepackage{booktabs}       % professional-quality tables
\usepackage{float}          % for H placement specifier
\usepackage{graphicx}       % graphics
\usepackage{amsfonts}       % blackboard math symbols
\usepackage{amsmath}        % math
\usepackage{nicefrac}       % compact symbols for 1/2, etc.
\usepackage{microtype}      % microtypography
\usepackage{inconsolata}    % better monospace font
\usepackage{xcolor}         % colors
\usepackage{colortbl}       % colored table cells
\usepackage{caption}        % caption spacing
\usepackage{enumitem}       % itemize spacing
\usepackage{lipsum}         % for dummy text
\usepackage{comment}        % for commenting out blocks
\usepackage{tabularx}       % tables with text wrapping
\usepackage{arydshln}       % dashed lines in tables
\usepackage{latexsym}       % for \textquotesingle
\usepackage{multirow}       % multirow in tables
\usepackage{tcolorbox}      % for highlighted boxes
\tcbuselibrary{skins}       % required for `enhanced` key used in promptbox
\usepackage{xspace}
\usepackage{hyperref}       % hyperlinks - must be loaded after acl
\definecolor{linkcolor}{RGB}{0, 0, 128}
\hypersetup{
    colorlinks   = true,
    citecolor    = linkcolor,
    linkcolor    = linkcolor,
    urlcolor     = linkcolor,
}
\setlist{parsep=0pt}

\newlength{\listtopsep}
\newlength{\listitemsep}
\SetEnumitemKey{complist}{leftmargin=5mm}

\definecolor{green}{HTML}{4DB78C}
\definecolor{cherry}{HTML}{CD3572}
\definecolor{lightcrest}{HTML}{FFE2C8}
\definecolor{cambridgeblue}{HTML}{8EE8D8}
\definecolor{cambridgelightblue}{HTML}{D1F9F1}
\newcommand{\appref}[1]{%
  \hyperref[#1]{Appendix~\ref*{#1}}%
}

\newcommand{\methodname}{\textsc{Polyglot Score}}
\newcommand{\shortmethodname}{\textsc{PG-Score}}

\newcommand{\teacher}{T}
\newcommand{\student}{S_{T,\ell}}
\newcommand{\basestudent}{S_{\text{base}}}
\newcommand{\refstudent}{S_{\text{ref}}}
\newcommand{\gendata}{\mathcal{X}_{T,\ell}}
\newcommand{\seeddata}{\mathcal{D}_{\text{seed},\ell}}

\newsavebox{\sbsTbl}
\newsavebox{\sbsFig}

\newlength{\sbsW}
\newcommand{\numlanguages}{6}
\newcommand{\nummodels}{10}
\newcommand{\numstudents}{240}
\newcommand{\basemodel}{OLMo 3 7B}

\definecolor{promptboxborder}{HTML}{254eff}
\definecolor{promptboxbody}{HTML}{f8ecda}
\newcommand{\promptbox}[2][]{%
    \begin{tcolorbox}[
            enhanced,
            colback=promptboxbody,
            colframe=promptboxborder,
            colbacktitle=promptboxborder,
            coltitle=white,
            fonttitle=\sffamily\small,
            boxrule=0.6pt,
            arc=2pt,
            left=6pt, right=6pt, top=4pt, bottom=4pt,
            title=#1,
        ]
        #2
    \end{tcolorbox}%
}

\newcommand{\huggingface}{\raisebox{-1.5pt}{\includegraphics[height=1.05em]{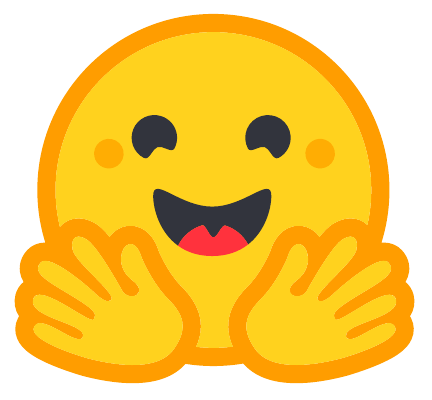}}\xspace}
\newcommand{\github}{\raisebox{-1.5pt}{\includegraphics[height=1.05em]{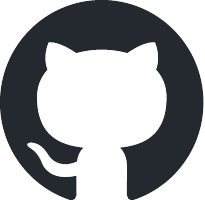}}\xspace}
\definecolor{chiablue}{HTML}{254EFF}

\title{Polyglot Teachers: Evaluating Language Models for Multilingual \\Synthetic Data Generation}

\author{%
Lester James V. Miranda\quad
\bf Ivan Vuli\'c\quad
\bf Anna Korhonen\quad \\
Language Technology Lab, University of Cambridge\\
\texttt{ljvm2@cam.ac.uk}\\\\
{\small\huggingface{}~~\textbf{Collection}\hspace{0.5em}\href{https://huggingface.co/collections/ljvmiranda921/polyglot-teachers}{\texttt{ljvmiranda921/polyglot-teachers}}}\quad
{\small\github{}~~\textbf{Code}\hspace{0.5em}\href{https://github.com/ljvmiranda921/polyglot-teachers}{\texttt{ljvmiranda921/polyglot-teachers}}}
}

\begin{document}

\maketitle

\begin{abstract}
    Synthesizing supervised finetuning (SFT) data from language models (LMs) to teach smaller models multilingual tasks has become increasingly common.
    However, teacher model selection is often ad hoc, typically defaulting to the largest available option,
    even though such models may have significant capability gaps in non-English languages.
    This practice can result in poor-quality synthetic data and suboptimal student downstream performance.
    In this work, we systematically characterize what makes an effective multilingual teacher.
    We combine intrinsic measures of data quality with extrinsic student model performance in a metric we call \methodname{}.
    We evaluate \nummodels{} LMs across \numlanguages{} typologically diverse languages,
    generating over 1.4M SFT examples and training \numstudents{} student models.
    Our analyses reveal that model scale alone does not significantly predict teacher effectiveness:
    the most effective teachers we identify are consistently smaller than the largest models evaluated, and their ranking is stable across student base model families.
    Instead, data qualities such as prompt diversity, length, and response fluency capture 93.3\% of the variance in intrinsic data quality and predict student performance.
    % Teacher effectiveness also correlates with language representation in pretraining corpora.
    Finally, we provide practical recommendations, including
    matching the model families of teacher-student pairs and generating responses to existing prompts or translating them from English, which can yield improvements for less-resourced languages.
    We hope that our work advances data-centric research in multilingual synthetic data and LM development.
\end{abstract}

\addtocontents{toc}{\protect\setcounter{tocdepth}{0}}
\section{Introduction}

\begin{figure}[t]
    \centering
    \includegraphics[width=0.95\linewidth, trim={0 0 0 0}]{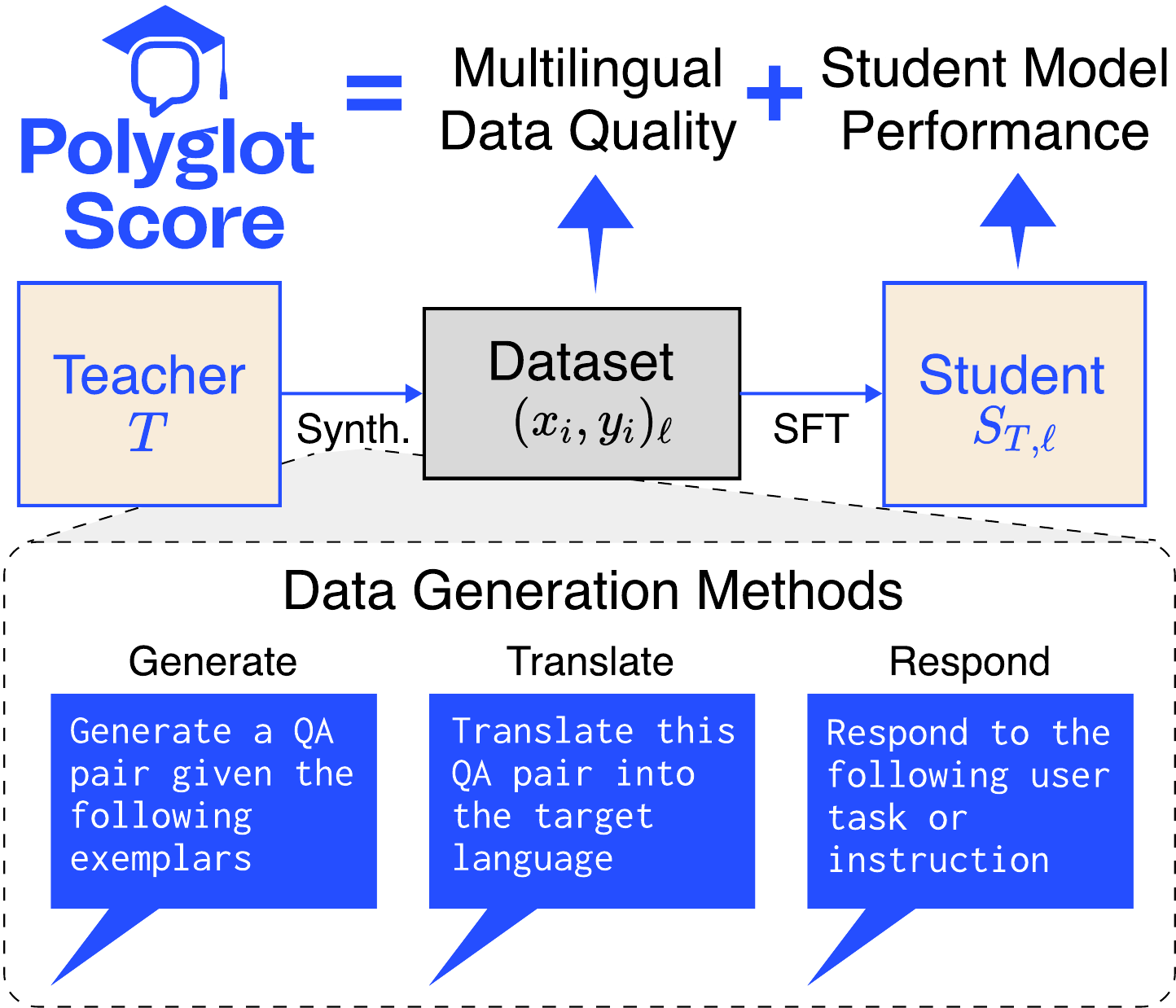}
    \caption{
        \textbf{
            Systematic evaluation of language models as multilingual teachers.
            % Overview of our method for evaluating language models as multilingual teachers (\methodname{}).
        }
        Across a synthetic data pipeline for language $\ell$,
        we evaluate a teacher model $\teacher$ on their ability to generate supervised finetuning data $(x_i,y_i)_{\ell}$ in order to train a capable student model $\student$.
        We examine three data generation methods: generate, translate, and respond.
        %\textit{Generate} a prompt-response pair given few-shot examples, \textit{Translate} prompts from English and generate a response, and \textit{Respond} to a prompt in the target language.
        Our metric, \methodname{}, incorporates both intrinsic data quality metrics and extrinsic student model performance to assess the effectiveness of a teacher model for a target language.
    }
    \label{fig:main_figure}
\end{figure}

Supervised finetuning \citep[SFT,][]{ouyang2022traininglanguagemodelsfollow} has emerged as a standard approach for adapting language models (LMs) to specific target languages \citep[\textit{inter alia}]{aryabumi2024aya,shengyu2025instruction}.
Central to the success of SFT is the availability of high-quality training data, consisting of pairs of user prompts and corresponding responses, which is often scarce for less-resourced languages \citep{kunchukuttan-etal-2025-data}.
Generating prompt-response pairs for these languages demands substantial human effort \citep{singh-etal-2024-aya,kapania-etal-2025-examining}, creating a bottleneck for language-specific model development.

To alleviate the challenge of human effort and data scarcity, synthetic data generation using LMs has gained traction as a promising solution for multilingual LM development \citep[\textit{inter alia}]{cahyawijaya-etal-2024-cendol,ng2025sea, martins2025eurollm,hammoud2025hala}.
This approach involves leveraging a typically larger \textbf{teacher model} to generate training examples, which are then used to finetune a smaller \textbf{student model} to replicate the knowledge of the teacher \citep{kim-rush-2016-sequence}.
However, existing works often select teacher models arbitrarily, defaulting to the largest state-of-the-art models that excel on benchmarks \citep{xu-etal-2025-stronger,li-etal-2025-small-models,zhang2025find}.
This practice is problematic because these models, despite strong performance, may have significant capability gaps in non-English languages, leading to poor-quality synthetic data that propagates the teacher's weaknesses rather than its strengths.
We therefore ask: \textit{``what makes an effective multilingual teacher for synthetic data generation, and how can we systematically measure it?''}

% In this work, we conduct a comprehensive analysis of \nummodels{} LMs across \numlanguages{} typologically diverse languages on three common synthetic data generation methods:
% responding to a user query or instruction,
% translating prompts from English to a target language,
% and
% generating prompt-response pairs given in-context examples (\S\ref{sec:intrinsic-metrics}).
% To systematically assess teacher model effectiveness, we introduce a framework called \textbf{\methodname{}} (\shortmethodname{}) that evaluates LMs using both \textbf{intrinsic measures of data quality} (\S\ref{sec:intrinsic-metrics}, i.e., the diversity of prompts and responses, the perplexity of the base model on the response, and response quality based on a multilingual reward model) and an \textbf{extrinsic measure of student model performance} on multilingual tasks (\S\ref{sec:extrinsic-metrics}, cultural understanding, mathematical reasoning, general chat).
% We aggregate these measurements into a single metric to provide a holistic assessment of a teacher model's data generation capabilities.

In this work, we conduct a comprehensive analysis of \nummodels{} LMs across \numlanguages{} typologically diverse languages on three common synthetic data generation methods:
responding to a user query or instruction,
translating prompts from English to a target language,
and
generating prompt-response pairs given in-context examples (\S\ref{sec:synthetic-data-generation}).
To systematically assess teacher model effectiveness,
we evaluate LMs using both 
\textbf{intrinsic measures of data quality} (\S\ref{sec:intrinsic-metrics}, i.e., the diversity of prompts and responses, the perplexity of the base model on the response, and response quality based on a multilingual reward model) 
and an \textbf{extrinsic measure of student model performance} on multilingual tasks (\S\ref{sec:extrinsic-metrics}, cultural understanding, mathematical reasoning, general chat).
We aggregate these measurements into a single metric called \methodname{} (\shortmethodname{}), in order to provide a holistic assessment of a teacher model's data generation capabilities.
Our contributions are as follows:

\begin{itemize}[complist]
    \item We close the \textbf{evaluation gap} by evaluating \nummodels{} teacher models, generating over 1.4M SFT examples and finetuning \numstudents{} student models from \basemodel{}.
        We find that Gemma 3 27B consistently ranks within the top three highest \shortmethodname{}
        and that the Gemma 3 model family outperforms other families such as Llama 3.1 and IBM Granite (\S\ref{sec:sota_teachers}).
        Our \shortmethodname{} rankings are consistent across other base model families (Llama 3.1 8B, Qwen 3 8B, Gemma 3 4B, \S\ref{sec:generalization_base_models}).
        % and model scale (7B, 32B, \appref{appendix:model_scale}).
    \item We provide \textbf{analyses and insights} on the characteristics of a good multilingual teacher model.
        Our analyses reveal that model scale and benchmark performance, which are common assumptions about a ``strong'' model, do not significantly predict teacher effectiveness (\S\ref{sec:common_assumptions}).
        Instead, we find that qualities of the generated data, namely prompt diversity and length coupled with fluent and diverse responses, capture 93.3\% of the variance in intrinsic data quality metrics, and their principal components predict student performance with $R^2=0.664$ (\S\ref{sec:factors_pgscore}).
        % In addition, we observe a suggestive positive correlation between teacher effectiveness and language representation in pretraining corpora such as CommonCrawl (Spearman $\rho=0.886$, $p<0.05$) (\S\ref{sec:language}).
        % For example, German and Spanish have higher \shortmethodname{}s compared to other languages we tested.
    \item Based on these findings, we \textbf{recommend a recipe} (\S\ref{sec:discussion}) for generating multilingual synthetic data.
        For example, we find that matching the model families of the teacher and student is a useful default when teacher quality is unknown (\S\ref{sec:generalization_base_models}),
        and
        generating responses to existing prompts or translating from English can yield substantial improvements on less-resourced languages compared to a random mix of data generation methods, though gains vary by teacher model (\S\ref{sec:effect_generation_method}).
        We apply the full recipe to a held-out language, Tagalog, and show that each recommendation yields an observable gain on a language-specific benchmark (\S\ref{sec:case_study}).
\end{itemize}

We hope that this work paves the way for developing inclusive and equitable language technologies through high-quality and cost-effective data.
\makeatletter
\ifacl@finalcopy
    We release our code, data, and models to drive research in multilingual synthetic data generation.
\else
    % use this
    We will release our code, data, and models to drive research in multilingual synthetic data generation after the review period.
\fi
\makeatother

\section{Evaluating Language Models as Multilingual Teachers}
\label{sec:method}

% In order to assess the effectiveness of LMs as multilingual teachers in a single metric, we measure its extrinsic and intrisic\methodname{}.
The \methodname{} (\autoref{fig:main_figure}) of a teacher model $\teacher$ for a target language $\ell$ is based on the intrinsic quality of the synthetic data generated by the teacher (\S\ref{sec:intrinsic-metrics}) and the extrinsic performance of a student model $\student$ finetuned on this data (\S\ref{sec:extrinsic-metrics}).

\subsection{Seed Dataset Construction}
%Most multilingual synthetic pipelines rely on a seed dataset\textemdash either human-annotated, web-crawled, or automatically translated\textemdash to bootstrap the data generation process \citep[\textit{inter alia}]{muennighoff-etal-2023-crosslingual,cahyawijaya-etal-2024-cendol,nguyen-etal-2024-democratizing}.
In order to bootstrap the synthetic data generation process, we create a seed dataset $\seeddata$ for each target language $\ell$.
We create $\seeddata$ by aggregating publicly available multilingual instruction-tuning datasets, including the Aya Collection \citep{aryabumi2024aya}, WildChat 4.8M \citep{zhao2024wildchat}, EuroBlocks-SFT \citep{martins2025eurollm}, and Magpie-Align \citep{xu2024magpie}.
In order to simulate scenarios where English prompts are translated into a target language, we also include examples from T\"ulu 3 SFT \citep{lambert2025tulu}, Helpsteer3 \citep[chosen responses,][]{wang2025helpsteer3humanannotatedfeedbackedit}, and GSM8K \citep[train split,][]{cobbe2021gsm8k}.
Detailed seed dataset statistics are in \appref{appendix:seed_dataset_statistics}.

\subsection{Synthetic Data Generation}
\label{sec:synthetic-data-generation}
Given a teacher model $\teacher$, target language $\ell$, and a seed dataset for language $\ell$, $\seeddata$,
we distill a synthetic dataset $\gendata = \{(x_i, y_i)\}_{i=1}^{N}$ consisting of $N$ prompt-response pairs $(x_i, y_i)$,
then finetune a base model $\basestudent$ on this data to obtain the student model $\student$:
\begin{equation}
    \begin{split}
        \gendata & = \teacher(\seeddata)                   \\
        \student & = \text{SFT}(\basestudent, \gendata)
    \end{split}
    \label{eq:pipeline}
\end{equation}

We consider three synthetic data generation methods found in literature (see \appref{appendix:lit-review-multilingual-sdg} for a brief review of multilingual synthetic data generation methods):
% (1) \textbf{generate}, where we sample $k$ prompt-response pairs from $\seeddata$ as few-shot examples and use $\teacher$ to generate a new pair $(x_i, y_i)$ conditioned on these examples \citep{rosenbaum-etal-2022-clasp,namboori2024gemquad,agrawal-etal-2023-qameleon},
% (2) \textbf{translate}, where we forward-translate English prompts from $\seeddata$ to the target language $\ell$ to obtain $x_i$, and use $\teacher$ to generate the corresponding response $y_i$ \citep{li2023bactrian,chitale2025role,de-gibert-etal-2025-scaling}, and
% (3) \textbf{respond}, where we take a prompt $x_i$ from $\seeddata$ and use $\teacher$ to generate the corresponding response $y_i$ \citep{whitehouse-etal-2023-llm,nguyen-etal-2024-democratizing}.

\begin{itemize}[complist]
    \item Generate: we sample $k$ prompt-response pairs from $\seeddata$ as few-shot examples and use $\teacher$ to generate a new pair $(x_i, y_i)$ conditioned on these examples.% \citep{rosenbaum-etal-2022-clasp,namboori2024gemquad,agrawal-etal-2023-qameleon}.
    \item Translate: we forward-translate English prompts from $\seeddata$ to the target language $\ell$ to obtain $x_i$, and use $\teacher$ to generate the corresponding response $y_i$.% \citep{li2023bactrian,chitale2025role,de-gibert-etal-2025-scaling}.
    \item Respond: we take a prompt $x_i$ from $\seeddata$ and use $\teacher$ to generate the response $y_i$.% \citep{whitehouse-etal-2023-llm,nguyen-etal-2024-democratizing}.
\end{itemize}

% We provide a brief review of multilingual synthetic data generation methods in \S\ref{sec:related_work} and a supplementary survey in \appref{appendix:lit-review-multilingual-sdg}.

\subsection{Intrinsic: Multilingual Data Quality \& Diversity Metrics}
\label{sec:intrinsic-metrics}

% \paragraph{Data quality and diversity metrics}
Data is valuable when it is both high-quality and diverse \citep{raventos2023pretraining,chen2024diversitysyntheticdataimpact,zhu2025bareleveragingbaselanguage}.\footnote{We use ``data quality'' to refer to both aspects hereafter.}
To estimate the value of $\gendata$, we compute a set of lexical and model-based metrics:

\begin{itemize}[complist]
    \item Diversity of prompts and responses $(d_x, d_y)$: a corpus-level statistic that computes the average pairwise cosine distance between the prompt and response embeddings.
          In practice, we use Llama-Embed-Nemotron-8B \citep{babakhin2025llama}, the top-performing model on the MMTEB leaderboard \citep{enevoldsen2025mmteb}, to embed the texts.
    \item Perplexity (PPL): the perplexity of a base model on the response $y_i$ conditioned on the prompt $x_i$, measuring the fluency and naturalness of the generated text. Lower perplexity indicates more coherent and linguistically natural responses.
    \item Reward score of a multilingual reward model (R): the verbalized score (1--5) of a multilingual reward model based on rubrics relating to fluency, naturalness, and instruction-following.
          In practice, we prompt M-Prometheus 14B \citep{pombal2025mprometheus} as an LM judge to score the quality of the prompt-response pair (\autoref{fig:llm_as_judge_prompt}).
          We choose M-Prometheus because of its high performance on human-aligned evaluation benchmarks, suggesting that the reward model aligns well with native speakers.
\end{itemize}

We combine these intrinsic metrics by scaling each metric using z-score normalization and averaging them as shown in \autoref{eq:intrinsic_score}.

\vspace{-1.5em}
\begin{equation}
    \begin{split}
        \text{Intrinsic}_{T,\ell} & = \frac{1}{\vert M \vert} \sum_{m \in M} \text{z-score}(m(\gendata)) \\
        \text{where } M           & = \{d_x, d_y, -\log(1+\text{PPL}), R\}
    \end{split}
    \label{eq:intrinsic_score}
\end{equation}

\definecolor{cambridgeblue}{HTML}{254EFF}
\definecolor{warmcrest}{HTML}{C96A2E}

\begin{table*}[t]
    \centering
    % \begin{noindent}
    %\renewcommand{\arraystretch}{0.9}
    \resizebox{\textwidth}{!}{%
    \begin{tabular}{@{}lw{c}{1.2cm}ccccccc@{}}
        \toprule
        \textbf{Teacher Model} & \textbf{Average} & \textbf{Arabic (ar)} & \textbf{Czech (cs)} & \textbf{German (de)} & \textbf{Spanish (es)} & \textbf{Indonesian (id)} & \textbf{Japanese (ja)} \\
        \midrule
        Gemma 3 27B Inst. & \cellcolor{cambridgeblue!12}0.726 & \cellcolor{cambridgeblue!2}\textbf{0.145} & \cellcolor{cambridgeblue!6}\textbf{0.360} & \cellcolor{cambridgeblue!27}1.655 & \cellcolor{cambridgeblue!22}\textbf{1.358} & \cellcolor{cambridgeblue!3}0.214 & \cellcolor{cambridgeblue!10}\underline{0.626} \\
        Aya Expanse 32B & \cellcolor{cambridgeblue!11}0.706 & \cellcolor{white}\underline{$-$0.058} & \cellcolor{cambridgeblue!3}0.222 & \cellcolor{cambridgeblue!24}1.468 & \cellcolor{cambridgeblue!18}1.129 & \cellcolor{cambridgeblue!19}\textbf{1.153} & \cellcolor{cambridgeblue!5}0.320 \\
        Gemma 3 12B Inst. & \cellcolor{cambridgeblue!9}0.595 & \cellcolor{warmcrest!7}$-$0.464 & \cellcolor{cambridgeblue!5}\underline{0.327} & \cellcolor{cambridgeblue!29}\underline{1.756} & \cellcolor{cambridgeblue!20}\underline{1.228} & \cellcolor{cambridgeblue!2}0.151 & \cellcolor{cambridgeblue!9}0.573 \\
        Command A & \cellcolor{cambridgeblue!9}0.546 & \cellcolor{warmcrest!22}$-$1.360 & \cellcolor{cambridgeblue!1}0.114 & \cellcolor{cambridgeblue!27}1.673 & \cellcolor{cambridgeblue!18}1.102 & \cellcolor{cambridgeblue!17}\underline{1.063} & \cellcolor{cambridgeblue!11}\textbf{0.683} \\
        Gemma 3 4B Inst. & \cellcolor{cambridgeblue!7}0.469 & \cellcolor{warmcrest!8}$-$0.488 & \cellcolor{cambridgeblue!5}0.330 & \cellcolor{cambridgeblue!27}1.644 & \cellcolor{cambridgeblue!15}0.929 & \cellcolor{warmcrest!1}$-$0.105 & \cellcolor{cambridgeblue!8}0.504 \\
        GPT-4o mini & \cellcolor{cambridgeblue!7}0.461 & \cellcolor{warmcrest!18}$-$1.117 & \cellcolor{white}0.015 & \cellcolor{cambridgeblue!29}\textbf{1.766} & \cellcolor{cambridgeblue!15}0.908 & \cellcolor{cambridgeblue!16}1.003 & \cellcolor{cambridgeblue!3}0.189 \\
        IBM Granite 4.0 & \cellcolor{cambridgeblue!5}0.312 & \cellcolor{warmcrest!1}$-$0.072 & \cellcolor{white}$-$0.031 & \cellcolor{cambridgeblue!16}1.000 & \cellcolor{cambridgeblue!12}0.734 & \cellcolor{warmcrest!1}$-$0.079 & \cellcolor{cambridgeblue!5}0.321 \\
        IBM Granite Micro & \cellcolor{cambridgeblue!5}0.304 & \cellcolor{warmcrest!4}$-$0.282 & \cellcolor{cambridgeblue!4}0.290 & \cellcolor{cambridgeblue!18}1.102 & \cellcolor{cambridgeblue!13}0.783 & \cellcolor{warmcrest!5}$-$0.329 & \cellcolor{cambridgeblue!4}0.264 \\
        Llama 3.1 70B Inst. & \cellcolor{cambridgeblue!2}0.140 & \cellcolor{warmcrest!16}$-$0.964 & \cellcolor{cambridgeblue!1}0.109 & \cellcolor{cambridgeblue!19}1.195 & \cellcolor{cambridgeblue!11}0.688 & \cellcolor{cambridgeblue!3}0.182 & \cellcolor{warmcrest!6}$-$0.373 \\
        Llama 3.1 8B Inst. & \cellcolor{warmcrest!5}$-$0.356 & \cellcolor{warmcrest!28}$-$1.693 & \cellcolor{warmcrest!16}$-$0.974 & \cellcolor{cambridgeblue!14}0.891 & \cellcolor{cambridgeblue!3}0.182 & \cellcolor{cambridgeblue!5}0.322 & \cellcolor{warmcrest!14}$-$0.863 \\
        \bottomrule
    \end{tabular}
    }%
    % \end{noindent}
    \caption{
        \textbf{Top models with the highest \shortmethodname{} (average across six languages).}
        We evaluate teacher models with varying size and model family on \numlanguages{} typologically diverse languages.
        For each language, we highlight the best model in \textbf{bold} and the second-best model with an \underline{underline}.
        Detailed results with standard errors are in \autoref{table:sota_evals_appendix}.
    }
    \label{table:sota_evals}
\end{table*}

\subsection{Extrinsic: Student Model Performance}
\label{sec:extrinsic-metrics}

We perform supervised finetuning of a base model $\basestudent$ on the synthetic dataset $\gendata$ to obtain a student model $\student$ (\autoref{eq:pipeline}).
Then, we evaluate $\student$ on a suite of multilingual tasks to assess how well the student has learned from the teacher.
These tasks include:

\begin{itemize}[complist]
    \item Cultural and factual understanding (\textsc{Culture}): we evaluate on Global-MMLU Lite \citep{singh-etal-2025-global}, containing culturally diverse and relevant questions that native speakers localized from English \citep{hendrycks2021measuring}.
    \item General chat (\textsc{Chat}): we evaluate on M-RewardBench \citep{gureja-etal-2025-rewardbench}, which measures the alignment of models with human preferences in conversational settings.
    \item Mathematical reasoning (\textsc{Math}): we evaluate on M-GSM \citep{shi2023language}, a multilingual version of the GSM8K dataset \citep{cobbe2021gsm8k} that tests the model's ability to solve mathematical word problems.
\end{itemize}

We choose these benchmarks because they cover distinct capabilities, are available across all target languages, and are widely adopted in the evaluation suites of recent multilingual LMs \citep[\textit{inter alia}]{grattafiori2024llama,team2025gemma,yang2025qwen3technicalreport}.
Instead of reporting the absolute score, we use the \textit{performance gap recovered} metric \citep[PGR,][]{kim-etal-2025-evaluating} 
to measure the improvement of a student over its base model relative to a reference model $\refstudent$ on the same scale.
% \vspace{-1em}
\begin{equation}
    \begin{split}
        \text{Extrinsic}_{T,\ell} & = \frac{1}{\vert B \vert} \sum_{b\in B} \text{PGR}_b(\student)                                                     \\
        \text{PGR}_b(\student)   & = \dfrac{\text{score}_b(\student) - \text{score}_b(\basestudent)}{\text{score}_b(\refstudent) - \text{score}_b(\basestudent)} \\
        \text{where } B           & = \{\textsc{Culture}, \textsc{Chat}, \textsc{Math}\}
    \end{split}
    \label{eq:extrinsic_score}
\end{equation}

Here, $\refstudent$ is an instruction-tuned model finetuned from the same pretrained checkpoint as $\basestudent$, and both are held fixed across all teachers.
For our main experiments, we use OLMo 3 7B as $\basestudent$ and OLMo 3 7B Instruct SFT as $\refstudent$.

\subsection{\methodname{} Computation}

To provide straightforward comparisons between teacher models, \shortmethodname{} reports a single score that combines both extrinsic and intrinsic metrics as shown in \autoref{eq:pg_score}.

\vspace{-1.5em}
\begin{equation}
    % \begin{split}
    \text{\shortmethodname{}}_{T,\ell} = \text{z-score}(\text{Intr.}_{T,\ell} + \text{Extr.}_{T,\ell})
    % \end{split}
    \label{eq:pg_score}
\end{equation}

We combine both intrinsic and extrinsic metrics because they capture complementary aspects of teacher quality.
Extrinsic metrics alone may overlook the quality of synthetic data that propagates through the ecosystem, while intrinsic metrics alone do not guarantee that the student model achieves strong downstream performance.
% For low-resource multilingual settings where synthetic data is scarce and costly to produce, evaluating both dimensions ensures that teacher models generate high-quality outputs that also translate to effective student learning.
The resulting \shortmethodname{} is z-score normalized, where $0$ indicates average teacher effectiveness, and higher scores indicate better synthetic data quality and student performance for that language.
We adopt equal weighting as a baseline; we show that
teacher rankings are robust to alternative weighting schemes in \appref{appendix:pgscore_ablation}.

\section{Experiments: Evaluating LMs and \shortmethodname{} Generalization}
\label{sec:experiments}

In this section, we measure the \methodname{} of state-of-the-art LMs (\S\ref{sec:sota_teachers}).
Then, we test whether our findings are consistent across other base models (\S\ref{sec:generalization_base_models}).
Finally, we determine if a certain data generation method is more effective in multilingual settings (\S\ref{sec:effect_generation_method}).
We conduct additional experiments and ablations in \appref{appendix:additional_experiments}.

\subsection{Which State-of-the-Art LMs Are Good Multilingual Teachers?}
\label{sec:sota_teachers}

\paragraph{Setup}
In order to evaluate the effectiveness of different LMs as multilingual teachers,
we select \nummodels{} state-of-the-art models that vary in scale, architecture, and training data,
then evaluate them on \numlanguages{} typologically diverse languages
by generating 10.5k prompt-response pairs for each teacher-language pair
where each data generation method (\S\ref{sec:synthetic-data-generation}) is equally represented.
We repeat the data generation process three times with different random seeds to account for variability in LM outputs.
Then, we finetune a pretrained \basemodel{} model \citep{olmo3-2025} on each $\gendata$ to obtain $\student$.
\appref{appendix:finetuning_details} describes SFT information.

\paragraph{Teacher Models}
We include Llama 3.1 \citep[8B, 70B,][]{grattafiori2024llama}, Gemma 3 \citep[4B, 12B, 27B,][]{team2025gemma}, Command A \citep{cohere2025command}, Aya Expanse 32B \citep{dang2024aya}, and IBM Granite \citep[4.0, Micro,][]{ibm-granite4.0-2025}.
%In addition, we also include closed-source models such as GPT-4o mini \citep{hurst2024gpt} and Claude \citep[Sonnet 4.5, Haiku 4.5,][]{claude3modelcard2024}.
In addition, we also include GPT-4o mini \citep{hurst2024gpt} as a representative closed-source model.
See \autoref{table:teacher_model_details} in \appref{appendix:teacher_target_details} for detailed model information.
% For each $\teacher$, we generate 10.5k prompt-response pairs for \numlanguages{} typologically diverse languages (Arabic, Czech, German, Spanish, Indonesian, and Japanese), where each data generation method (\S\ref{sec:intrinsic-metrics}) is equally represented.

\paragraph{Target Languages}
We select \numlanguages{} typologically diverse languages: Arabic (ar), Czech (cs), German (de), Spanish (es), Indonesian (id), and Japanese (ja).
These languages are chosen due to their variation in resource availability, script, and family.
This language choice is also supported by prior work on informed sampling \citep{ploeger-etal-2026-principled} that considers typological variety of the chosen languages.
See \autoref{table:language_details} in \appref{appendix:teacher_target_details} for language statistics.

\begin{figure*}[t]
    \centering
    % Two panels, each exactly one text column wide (\sbsW), separated by \columnsep.
    \begin{minipage}[b]{\sbsW}
        \centering
        \resizebox{\sbsW}{!}{%
        {\footnotesize
            \setlength{\tabcolsep}{2pt}
            \begin{tabular}{@{}lrrrr@{}}
                \toprule
                                       & \multicolumn{4}{c}{\textbf{Base Model} ($\basestudent$)}                                                                                        \\
                \cmidrule(lr){2-5}
                \textbf{Teacher Model} & \textbf{OLMo 3 7B}                                   & \textbf{Gemma 3 4B}        & \textbf{Qwen 3 8B}         & \textbf{Llama 3.1 8B}        \\
                \midrule
                GPT-4o mini            & 0.551                             & \cellcolor{chiablue!24}1.022 & 1.005                      & 0.621                      \\
                Llama 3.1 70B Inst.    & 0.138                                                & 0.338                      & 1.039                      & 0.497                      \\
                Llama 3.1 8B Inst.     & $-$0.160                                             & $-$0.133                   & 0.365                      & 0.048                      \\
                Command A              & 0.459                                                & 0.725                      & 0.974                      & 0.737                      \\
                Aya Expanse 32B        & \cellcolor{chiablue!24}0.854                           & \cellcolor{chiablue!6}0.762 & 1.183                      & 0.793                      \\
                Gemma 3 27B Inst.      & \cellcolor{chiablue!6}0.672                           & \cellcolor{chiablue!13}0.810   & \cellcolor{chiablue!6}1.301   & \cellcolor{chiablue!6}0.800 \\
                Gemma 3 12B Inst.      & 0.481                                                & 0.666                      & \cellcolor{chiablue!13}1.393 & \cellcolor{chiablue!13}0.804   \\
                Gemma 3 4B Inst.       & 0.350                                                & 0.712                      & 0.545                      & \cellcolor{chiablue!24}1.062 \\
                IBM Granite 4.0        & 0.283                                                & 0.278                      & 0.831                      & $-$0.001                   \\
                IBM Granite Micro      & 0.164                                                & 0.455                      & 1.079                      & 0.396                      \\
                Qwen 3 8B              & 0.398                                                & 0.615                      & \cellcolor{chiablue!24}1.421 & 0.662                      \\
                OLMo 3 7B Inst.        & \cellcolor{chiablue!13}0.731                           & 0.295                      & 0.581                      & 0.277                      \\
                \bottomrule
            \end{tabular}
        }%
        }%
        \captionsetup{type=table}
        \caption{
            \textbf{\shortmethodname{} of each teacher model across different base models
                (average across Arabic, German, and Indonesian).}
            We highlight the \colorbox{chiablue!24}{top}, \colorbox{chiablue!13}{second}, and
            \colorbox{chiablue!6}{third} best teacher models for each base model.
        }
        \label{tab:generalization_base_models}
    \end{minipage}%
    \hspace{\columnsep}%
    \begin{minipage}[b]{\sbsW}
        \centering
        \includegraphics[width=0.86\sbsW]{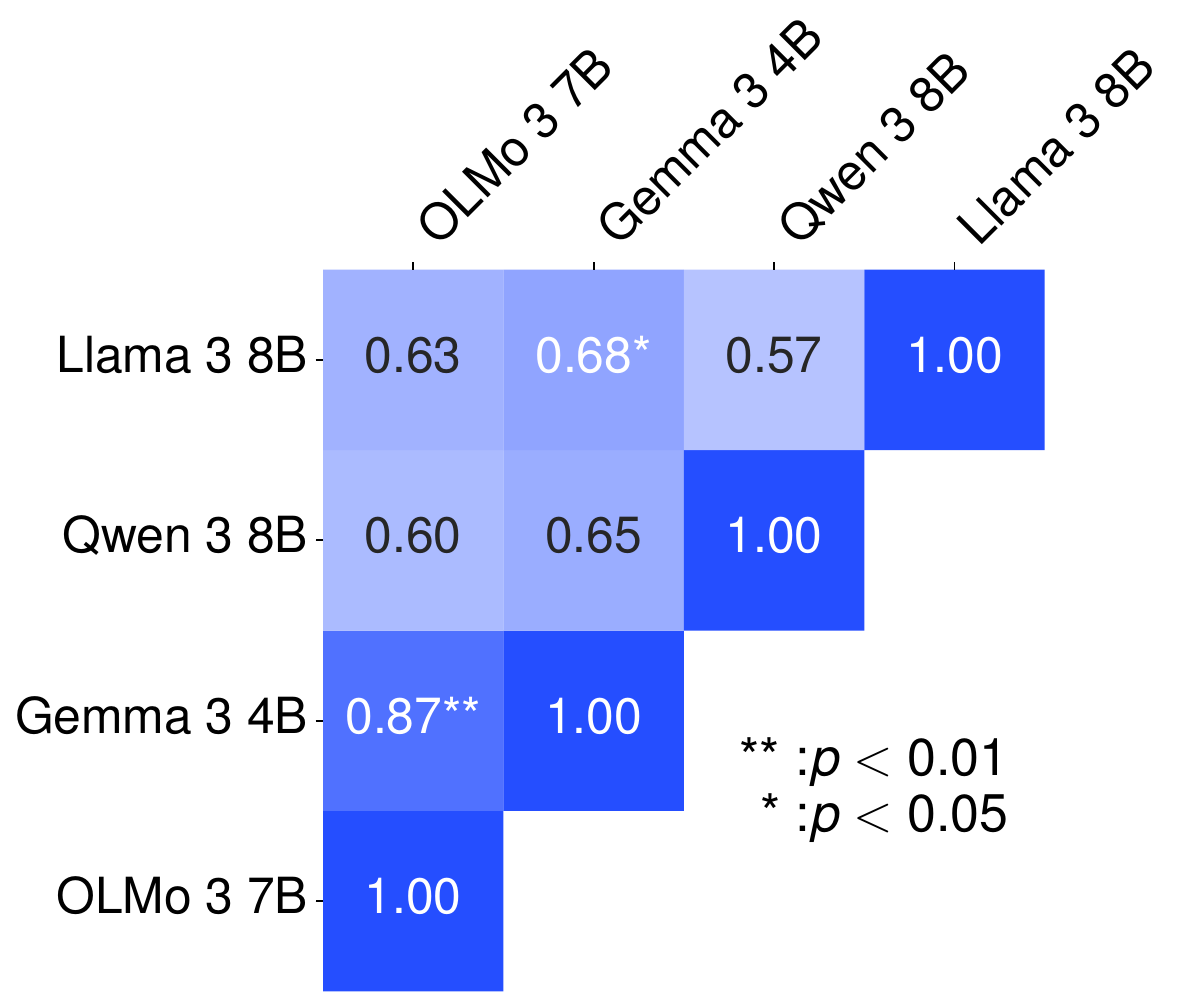}
        \captionsetup{type=figure}
        \caption{
            \textbf{Consistency of teacher rankings across base models.}
            Spearman rank correlation $\rho$ between the teacher rankings induced by each pair of
            base models.
        }
        \label{fig:generalization_base_models}
    \end{minipage}
\end{figure*}

% \begin{comment}
\begin{table*}[t]
    \centering
    \footnotesize
    \resizebox{\textwidth}{!}{%
        \setlength{\tabcolsep}{2pt}
        \begin{tabular}{@{}lrrrrrrrrr@{}}
            \toprule
             & \multicolumn{3}{c}{\textbf{Arabic (ar)}} & \multicolumn{3}{c}{\textbf{German (de)}} & \multicolumn{3}{c}{\textbf{Indonesian (id)}} \\
            \cmidrule(lr){2-4} \cmidrule(lr){5-7} \cmidrule(lr){8-10}
            % \begin{noindent}
            \textbf{Teacher Model} & Generate  & Translate  & Respond & Generate & Translate & Respond & Generate & Translate & Respond \\
            \midrule
            Gemma 3 27B Inst.      & 0.032  & 0.276  & \textbf{0.802}  & \textbf{2.140}    & 2.086     & 1.212   & 1.189    & \textbf{1.196}     & 0.046   \\
            Aya Expanse 32B        & $-$0.276  & \textbf{0.148}  & $-$1.349  & \textbf{1.473}    & 1.255     & 1.451  & 0.039    & 0.733     & \textbf{1.606}   \\
            Llama 3.1 70B Inst.    & $-$0.867 & $-$1.025  & $\mathbf{-}$\textbf{0.215}  & \textbf{1.391}    & 0.459     & 1.187  & $-$0.146    & 0.089     & \textbf{0.155} \\
            % \end{noindent}
            \bottomrule
        \end{tabular}%
    }
    \caption{
        \textbf{\shortmethodname{} across three data generation methods: Generate, Translate, and Respond (\S\ref{sec:synthetic-data-generation}).}
        For each data generation method, we generate 10k samples per teacher-language pair and finetune a student model on OLMo 3 7B.
        % We find that the \textit{Generate} method yields the highest \shortmethodname{} on German (high-resource), while the \textit{Respond} method is most effective on Arabic (medium-resource) and Indonesian (low to medium-resource).
        We show percentage increases in \shortmethodname{} compared to a baseline (equal representation of the three data generation methods) in \autoref{tab:pct_increase_data_generation_method}.
    }
    \label{table:data_generation_method}
\end{table*}

\paragraph{Results}
\autoref{table:sota_evals} shows the \shortmethodname{} of each teacher model across all target languages.
The results suggest the following:

\begin{itemize}[leftmargin=5mm]%,itemsep=2mm]%,topsep=0mm,itemsep=0mm]
    \item \textbf{Gemma 3 27B and Aya Expanse 32B are the most effective teachers.} Gemma 3 27B achieves the highest average \shortmethodname{} (0.726), followed closely by Aya Expanse 32B (0.706), both outperforming larger models such as Llama 3.1 70B Inst. (0.140), suggesting that model scale alone does not determine teacher effectiveness. We also observe that the Gemma 3 family dominates the top ranks, while the Llama 3.1 family underperforms on most languages.
    \item \textbf{Smaller LMs can be effective multilingual teachers.} Gemma 3 12B (0.595) and 4B (0.469) rank among the top-5 teachers, while the Llama 3.1 70B Inst. (0.140) ranks ninth, suggesting that smaller LMs can match or exceed larger LMs in data generation capabilities.
    \item \textbf{Teacher performance varies significantly by language.} German and Spanish consistently show the highest scores across all models, while Arabic proves challenging with most teachers yielding negative scores, suggesting that language-specific factors influence teacher effectiveness. We hypothesize that a language's resource status or presence in pretraining data may contribute to this variability (\appref{sec:language}).
          % \item \textbf{Model family matters more than scale.} The Gemma 3 family dominates the top ranks (27B, 12B, and 4B all in top-5), while both Llama 3.1 variants underperform despite the 70B model's size, suggesting that training data composition and instruction-tuning quality are critical factors for multilingual teaching effectiveness.
\end{itemize}

\subsection{Generalization of \shortmethodname{} Across Different Base Models}
\label{sec:generalization_base_models}

% We test whether the \methodname{} framework generalizes across different base models.

\paragraph{Setup}
Instead of using \basemodel{} as the base model ($\basestudent$) for student finetuning, we use (1) Llama 3.1 8B, (2) Gemma 3 4B PT, and (3) Qwen 3 8B Base \citep{yang2025qwen3technicalreport}.
We recompute $\basestudent$-dependent metrics such as perplexity and PGR.
To reduce computational costs, we focus on three languages: German (high \shortmethodname{}), Indonesian (mid-range), and Arabic (low \shortmethodname{}).

\paragraph{Results}
\autoref{tab:generalization_base_models} shows the average \shortmethodname{} of each teacher model across different base models.
We observe that the \textbf{best teacher models remain consistent across different student base models}, with Gemma 3 27B and Aya Expanse 32B consistently ranking among the top three teachers.
Furthermore, the Gemma 3 family continues to outperform other model families.
In addition, we find that the model rankings vary slightly depending on the base model used, as Spearman rank correlation ranges from $\rho=0.57$ (moderate) to $\rho=0.87$ (strong, \autoref{fig:generalization_base_models}).
We hypothesize that this variation may be due to differences in architecture and pretraining data between base models.
Despite this variation, we observe that \textbf{matching teacher and student model families is a useful default when teacher quality is unknown}.
The matched teacher ranks among the top two on three of the four base models we test (\autoref{tab:generalization_base_models}):
Gemma 3 teachers consistently perform well with Gemma 3 student bases,
while Qwen 3 8B is the strongest teacher on the Qwen 3 8B base and OLMo 3 7B Instruct ranks second on the OLMo 3 7B base.
The exception is Llama 3.1, which is a weak multilingual teacher overall (\S\ref{sec:sota_teachers}), suggesting that family matching helps only when the matched teacher is itself capable in the target languages.
This finding is reasonable given that models from the same family likely share similar tokenization schemes, leading to easier transfer from teacher to student.
% In addition, family-matching is not a hard constraint unlike in other distillation settings \citep[\textit{on-policy,}][]{agarwal2024onpolicydistillationlanguagemodels,boizard2025towards}.
% Gemma 3 teachers consistently perform well with Gemma 3 base models with a \shortmethodname{} increase of $+$20.53\% at minimum (i.e., for Gemma 3 27B Instruct teacher model and Gemma 3 4B base model).
% This suggests that while teacher selection exhibits some base model dependency, matching model families remains a practical strategy when the optimal teacher is unknown.
For our core experiment, we use \basemodel{} as the base model for finetuning
to control the effect of model family alignment when evaluating teacher quality.
% and
% (2) enable transparent analysis of its publicly available pretraining and post-training datasets.

\begin{table}[b]
    \centering
    %    \resizebox{\columnwidth}{!}{%
    \begin{tabular}{lccc}
        \toprule
        \textbf{Predictor}                    & $\mathbf{\beta}$ & \textbf{SE} & \textbf{p} \\
        \midrule
        $\log\left(\text{Param. Size}\right)$ & 0.053            & 0.080       & 0.507      \\
        Avg. Multilingual Perf.               & 1.387            & 2.204       & 0.529      \\
        \bottomrule
    \end{tabular}%
    %    }
    \caption{
        \textbf{Results from a mixed-effects regression model on \shortmethodname{} on an LM's (a) size and (b) avg. multilingual benchmark performance.}
        The lack of significant correlation suggests that both predictors are not solely sufficient to ensure teacher effectiveness.
    }
    \label{table:correlation}
\end{table}

\subsection{Effect of Synthetic Data Generation Method on \shortmethodname{}}
\label{sec:effect_generation_method}

\paragraph{Setup}
In order to determine if a data generation method is more effective than others,
we generate 10k prompt-response pairs for each method in \S\ref{sec:synthetic-data-generation} and compare the \shortmethodname{} of each mix.
We recompute intrinsic data quality metrics and finetune \basemodel{} to obtain a student model and evaluate the teacher's \shortmethodname{}.
We also compare each mix against a baseline consisting of 10k instances with roughly equal numbers of samples ($\approx$3.3k) from each method.
To reduce computational costs, we conduct this experiment on three representative teachers (Gemma 3 27B, Aya Expanse 32B, and Llama 3.1 70B) spanning high to low \shortmethodname{}, and three languages (German, Indonesian, Arabic) covering diverse resource levels.

\paragraph{Results}
\autoref{table:data_generation_method} shows the \shortmethodname{} of each data generation method (see \autoref{tab:pct_increase_data_generation_method} for baseline comparisons).
We observe that for a \textbf{high-resource language like German, the \textit{Generate} method yields the highest \shortmethodname{}}, while for \textbf{less-resourced languages like Arabic and Indonesian, the \textit{Respond} or \textit{Translate} methods are more effective}.
We hypothesize that this occurs because the \textit{Generate} method depends on few-shot examples from the seed dataset, which are typically of higher quality in high-resource languages.
Overall, our findings suggest that selecting a data generation method can have an impact on teacher effectiveness.
In our core experiment, we sample an equal mix of all three methods (3.5k each) to control their effect when evaluating teacher model quality.

\section{Analysis: What Makes a Good Multilingual Teacher Model?}
\label{sec:analysis}

We investigate the factors that contribute to effective multilingual teachers.
We start by analyzing common assumptions about teacher model performance, such as size and benchmark scores (\S\ref{sec:common_assumptions}),
then determine which intrinsic factors drive student performance (\S\ref{sec:factors_pgscore}).
% We perform additional analyses in the Appendix, concerning language properties (\appref{sec:language}), translation capabilities
% Lastly, in \appref{sec:language}, we examine language properties that might influence a teacher's \shortmethodname{}.
% Finally, we explore which models are the most cost-effective teachers (\S\ref{sec:cost_effective_teachers}).

\begin{table}[t]
    \centering
    % \resizebox{\columnwidth}{!}{%
    \begin{tabular}{lrr}
        \toprule
        \textbf{PC} & \textbf{Variance Expl.} & \textbf{Cumulative} \\
        \midrule
        PC1         & 42.2\%                  & 42.2\%              \\
        PC2         & 22.1\%                  & 64.3\%              \\
        PC3         & 16.5\%                  & 80.8\%              \\
        PC4         & 12.6\%                  & 93.3\%              \\
        \hdashline
        PC5         & 3.5\%                   & 96.8\%              \\
        PC6         & 3.2\%                   & 100.0\%             \\
        \bottomrule
    \end{tabular}%
    % }
    \caption{
        \textbf{Variance explained by principal components from intrinsic data quality metrics.}
        There are four principal components that explain 93.3\% (cumulative) of the variance.
    }
    \label{table:pca_results}
\end{table}

\subsection{Do stronger models make better teachers?}
\label{sec:common_assumptions}

\paragraph{Setup}
In order to determine if there is a relationship between a model's size or benchmark performance (i.e., common assumptions to assess a model's ``strength'') to its effectiveness as a multilingual teacher,
we fit a mixed-effects model regressing \shortmethodname{} on (a) parameter size (N=27, i.e., 9 models $\times$ 3 trials, excluding GPT-4o mini whose size is undisclosed), and (b) average multilingual benchmark performance on Global-MMLU Lite, M-GSM, and M-RewardBench (N=180, \nummodels{} models $\times$ \numlanguages{} languages $\times$ 3 trials).

\paragraph{Results}
\autoref{table:correlation} shows the regression results.
We observe that \textbf{neither parameter size nor average multilingual benchmark performance significantly predict \shortmethodname{}} ($p>0.05$).
Specifically, a 1-unit increase in $\log(\text{Param. Size})$ corresponds to a non-significant 0.053 increase in \shortmethodname{}.
Although this finding confirms the results of \citet{xu-etal-2025-stronger} and \citet{kim-etal-2025-evaluating} for English-based tasks,
we show that ``stronger'' models do not necessarily make better multilingual teachers.
% This finding challenges the common assumption that larger models or those with strong benchmark performance are inherently better teachers.

\subsection{Which intrinsic metrics determine extrinsic student model performance?}
\label{sec:factors_pgscore}

\begin{figure}[t]
    \centering
    \includegraphics[width=0.48\textwidth, trim={0cm 0cm 0cm 0cm}]{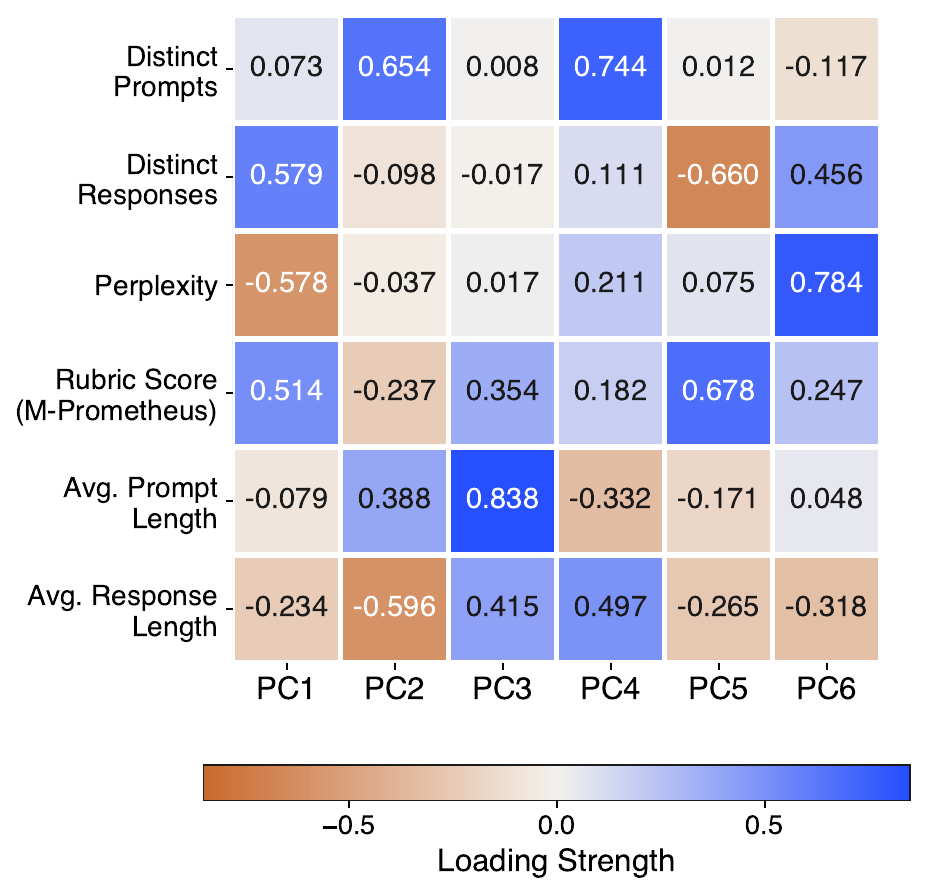}
    \caption{
        \textbf{Loading strength of intrinsic metrics on the principal components (PCs).}
        PC1 suggests that good teachers produce diverse and high-quality responses, while PC2 focuses on prompt diversity and length.
        PC3 and PC4, together, indicate the importance of prompts on student performance.
        % PC1 captures variation in response characteristics such as lower perplexity and high distinctiveness, while PC2 captures prompt diversity and length.
        % Higher absolute loading values indicate stronger contributions of each metric to the corresponding principal component.
    }
    \label{fig:pca_loading_factors}
\end{figure}

\begin{figure}[t]
    \centering
    \includegraphics[width=0.43\textwidth, trim={0cm 0cm 0cm 0cm}]{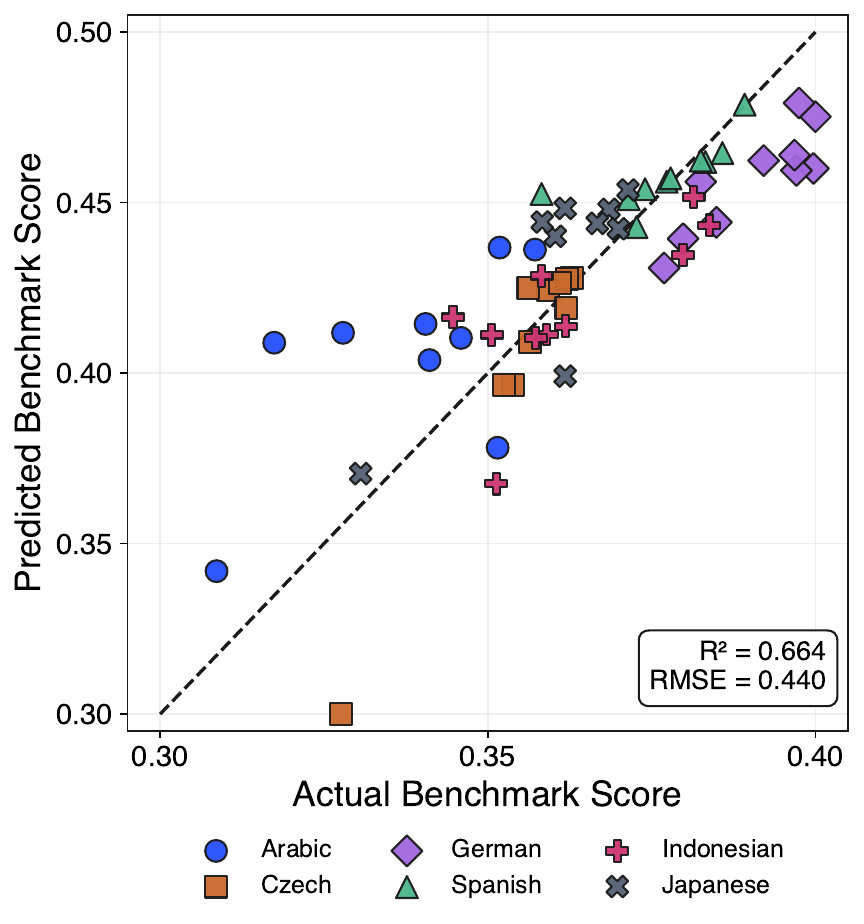}
    \caption{
        \textbf{Fit of a linear regression model on the PCs of the intrinsic metrics to predict student performance.}
        Intrinsic metrics, via their PCs, can predict extrinsic student performance ($R^2=0.664$ and $\text{RMSE}=0.440$) on multilingual benchmarks (\S\ref{sec:extrinsic-metrics}).
    }
    \label{fig:pca_regression}
\end{figure}

% \noindent We design \shortmethodname{} to combine both intrinsic and extrinsic metrics to holistically represent teacher model effeciveness.
% However, we hypothesize that downstream student performance can be predicted from intrinsic data quality metrics alone.
%\noindent Although \shortmethodname{} combines both intrinsic and extrinsic metrics, we are interested in understanding which intrinsic data quality metrics contribute to student performance.

\paragraph{Setup}
In order to identify latent factors from the intrinsic metrics that explain student performance, we perform principal component analysis (PCA) on the intrinsic metrics described in \S\ref{sec:intrinsic-metrics}.
Then, we fit a regression model to predict extrinsic student performance based on the principal components (PCs) obtained from PCA:
we split 180 data points (\nummodels{} models $\times$ \numlanguages{} languages $\times$ 3 trials) into 80\% train and 20\% test, then train a linear regression model with the PCs as the features and the student performance as the target.

\paragraph{Results}
\autoref{table:pca_results} shows how much of the variance is explained by each principal component while \autoref{fig:pca_loading_factors} shows the loading strength of each intrinsic metric on the principal components.
We observe that the first four PCs explain 93.3\% of the variance in the intrinsic data quality metrics.
Specifically, PC1 (42.2\%) captures characteristics such as \textbf{lower response perplexity and high distinctiveness}, PC2 (22.1\%) captures variance in characteristics such as \textbf{higher prompt diversity and length},
whereas PC3 (16.5\%) and PC4 (12.6\%) capture variance that reinforces trends on prompt length and diversity.
In addition, \autoref{fig:pca_regression} shows the fit of a linear model on the test set when the PCs learn to predict student performance.
We observe that interactions within the intrinsic metrics can predict extrinsic student performance reasonably well, with $R^2=0.664$ and $\text{RMSE}=0.440$.
This finding suggests that even with a simple linear model, our chosen \textbf{intrinsic metrics are predictive of student performance.}
In practice, these insights can help practitioners select teacher models based on intrinsic metrics alone, which are cheaper to compute than extrinsic student evaluations.

\section{Discussion: Towards a Recipe for Multilingual Synthetic Data Generation}
\label{sec:discussion}

Our results provide actionable insights for selecting and effectively using teacher models in multilingual synthetic data generation.
First, we find that \textbf{model scale does not significantly predict teacher effectiveness}:
Llama 3.1 70B Instruct, despite being one of the largest models evaluated, ranks at the bottom half in \shortmethodname{} across all student base models we tested (\S\ref{sec:sota_teachers}, \S\ref{sec:generalization_base_models}).
%Our findings align with prior work in English \citep{xu-etal-2025-stronger, kim-etal-2025-evaluating} and extend them to multilingual contexts.
Our analyses suggest that what matters instead is the quality of generated data:
prompt diversity, response fluency, and length collectively capture over 93\% of the variance in intrinsic data quality and predict student performance with $R^2=0.664$ (\S\ref{sec:factors_pgscore}), offering practitioners a cheaper alternative to full student training runs for screening teachers.

Second, when the optimal teacher is unknown, \textbf{matching model families offers a useful default for teacher selection}.
Gemma, Qwen, and OLMo teachers paired with students from their own family rank among the top two teachers on that base (\autoref{tab:generalization_base_models}), though this holds only when the matched teacher is itself a capable multilingual model.
We hypothesize this finding reflects shared tokenization and similar pretraining distributions, though disentangling these factors remains future work.

Finally, we find that \textbf{there are language-dependent considerations for data generation.}
For high-resource languages like German, where seed data quality is high, the \textit{Generate} method performs best.
For less-resourced languages like Arabic and Indonesian, methods that leverage existing prompts (\textit{Respond}) or transfer from English (\textit{Translate}) can yield substantial gains over a uniform mix of methods, though the magnitude varies by teacher (\autoref{table:data_generation_method}).
% However, even the \textbf{optimal data generation method cannot fully compensate for gaps in pretraining data}:
% teacher effectiveness shows a suggestive correlation with CommonCrawl representation ($\rho=0.886$, $N=6$ languages), and Arabic proves challenging for all teachers despite the best generation method.
For truly low-resource languages, we recommend combining synthetic data generation with targeted data collection.

\begin{figure*}[t]
    \centering
    \includegraphics[width=0.85\linewidth]{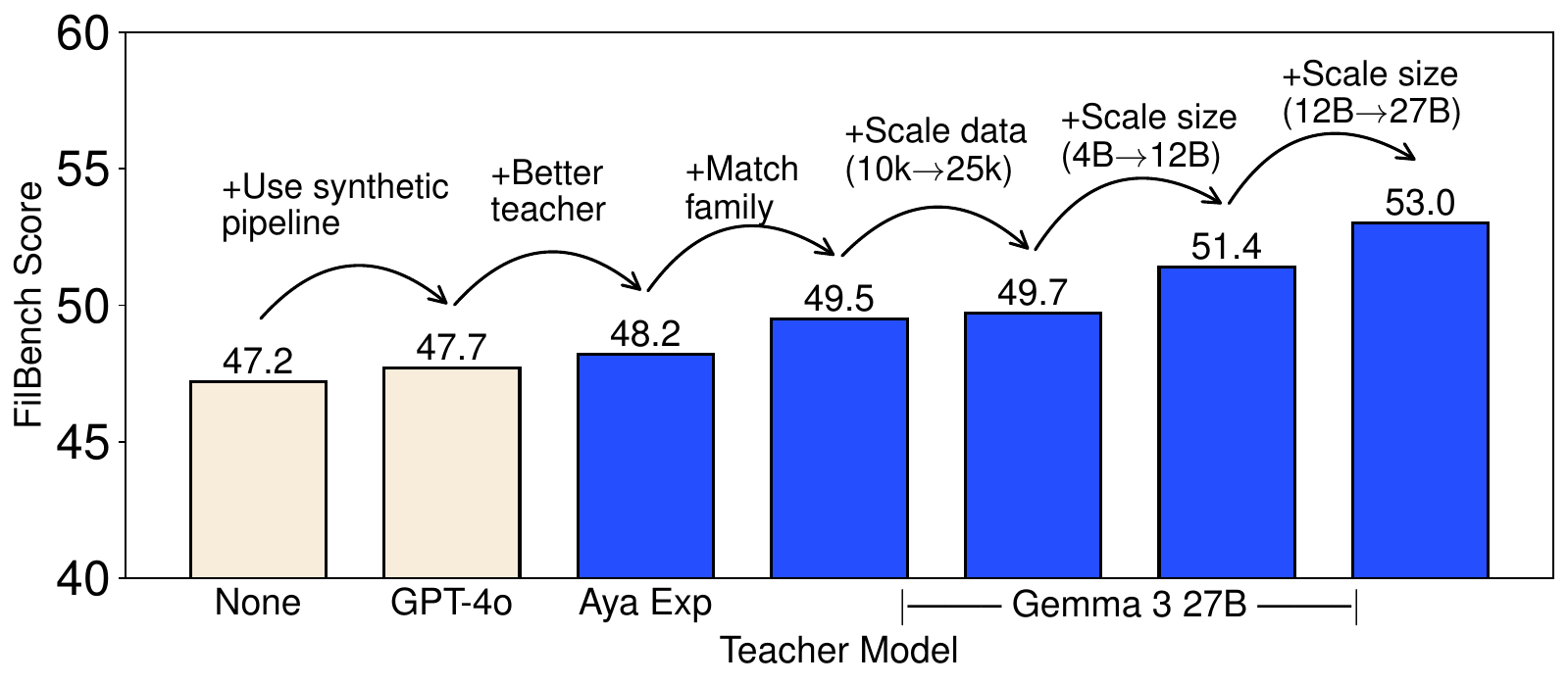}
    \caption{
        \textbf{Student model performance on a held-out language, Tagalog, across several synthetic data interventions.}
        Given a held-out language (Tagalog) and an evaluation benchmark (\textsc{FilBench}), we apply data interventions based on our recommendations for creating a multilingual synthetic data recipe (\S\ref{sec:discussion}).
        We provide ablation details and leaderboard results in \appref{appendix:tagalog_case_study}.
        %Interventions are additive, and we report \textsc{FilBench} scores in \appref{appendix:case_study_leaderboard}.
    }
    \label{fig:tgl_ablation_filbench_scores}
\end{figure*}

\section{Case Study: Applying the Synthetic Data Recipe to a Held-Out Language}
\label{sec:case_study}

As an application of our findings and discussion in \S\ref{sec:discussion},
we present a case study on developing a multilingual synthetic data recipe on a held-out language: Tagalog.
It is a mid-resource language (Category 3 in \citet{joshi-etal-2020-state}'s taxonomy) and the standardized form of Filipino, the national language of the Philippines.
We evaluate on \textsc{FilBench} \citep{miranda-etal-2025-filbench}, a Filipino-centric benchmark that was not used to compute \shortmethodname{}.
In order to measure the contribution of our findings and recommendations, we perform the following ablation experiments as shown in \autoref{fig:tgl_ablation_filbench_scores}.
Note that the interventions described below are additive, and we describe the full recipe in \appref{appendix:case_study_setup}.

We find that each intervention yields an observable gain.
Replacing publicly available Tagalog SFT data with synthetic instances from a GPT-4o mini teacher performs similarly ($\Delta=0.43$\,pp), suggesting that there is no significant advantage to using a synthetic data pipeline if the teacher model is not optimal.
Swapping in a teacher with a higher \shortmethodname{} (Aya Expanse 32B, 0.706 vs.\ 0.461) improves performance, suggesting that the metric is generalizable across an unseen language,
while matching the teacher to the family of the Gemma 3 4B base model yields a substantial further improvement.
Increasing data scale (10k to 25k instances, $+0.21$\,pp) and student model scale (4B to 12B and 27B) provide additional but smaller gains.
We report the full ablation, the \textsc{FilBench} leaderboard comparison, and a native-speaker inspection of the generated data in \appref{appendix:tagalog_case_study}.

\section{Related Work}
\label{sec:related_work}

\paragraph{Synthetic Data Generation for Multilingual SFT}
In order to offset the high costs of recruiting language experts for data collection, prior work has relied on generating synthetic datasets.
This effort resulted in large multilingual datasets such as Bactrian-X \citep[\textit{Translate,}][]{li2023bactrian}, MultiAlpaca \citep[\textit{Generate,}][]{wei2023polylmopensourcepolyglot}, and xP3 \citep[\textit{Respond,}][]{muennighoff-etal-2023-crosslingual} that were created through various data generation methods.
These works have different data generation recipes, and so 
we provide a brief survey of these works and their recipes in \appref{appendix:lit-review-multilingual-sdg}, 
then classify them across the three strategies (Generate, Translate, Respond; \S\ref{sec:synthetic-data-generation}).
% Recently, several works have explored building language-specific LMs using a combination of these techniques \citep{martins2025eurollm,sarvam2025sarvamm,ng2025sea} especially with the growing interest in training ``sovereign LMs'' \citep{bondarenko2025sovereignlargelanguagemodels,Dale_2025}. % that cater to the specific needs of a language community or nation.
Building on these prior efforts, we examine the three core strategies for multilingual synthetic data generation and test each in isolation.
This setup enables us to provide practitioners with an empirically grounded recipe for selecting teacher LMs that we hope is applicable across any generation method.

\paragraph{Evaluating and Improving the Synthetic Data Pipeline}
While prior works have evaluated aspects of the synthetic data pipeline, 
they typically do so in isolation (i.e., intrinsic \textit{or} extrinsic, but not both) or focus exclusively on English \citep{zhang2025find}.
For instance, \citet{kim-etal-2025-evaluating} evaluated teacher models solely as a function of \textit{extrinsic} student performance on English tasks (e.g., reasoning and coding), while \citet{cai2025opendataarenafairopenarena}'s OpenDataArena focuses on \textit{intrinsic} data quality (model-based and heuristic) to score models.
Signals of multilingual data quality are often a function of corpus-level diversity \citep{artetxe-schwenk-2019-margin,enevoldsen2025mmteb,sam2025analyzing} and generation quality \citep{pombal2025mprometheus,anugraha2026mr}.
On the other hand, multilingual LMs are typically evaluated on general-knowledge and culture-specific benchmarks \citep[\textit{inter alia}]{qin2025survey,team2025gemma,salamanca2026tinyayabridgingscale}.
These practices informed our choice of intrinsic and extrinsic metrics throughout this work.
More importantly, \shortmethodname{} provides a holistic metric that combines both intrinsic data quality and extrinsic student downstream performance to evaluate teacher models across various generation methods.

\section{Conclusion}

We present \methodname{}, a metric that evaluates LMs as multilingual teachers by combining the intrinsic quality of the synthetic data they generate with the extrinsic performance of the students they train.
Across \nummodels{} teachers, \numlanguages{} typologically diverse languages, and \numstudents{} student models, we find that model scale alone does not predict teacher effectiveness:
Gemma 3 27B and Aya Expanse 32B are the strongest teachers we evaluate, both outperforming Llama 3.1 70B Instruct.
What matters instead are properties of the generated data, as prompt diversity, length, and response fluency capture 93.3\% of the variance in our intrinsic metrics and predict student performance with $R^2=0.664$.
We distill these results into a recipe for multilingual synthetic data generation and validate it on a held-out language, Tagalog, where each recommendation yields an observable gain on \textsc{FilBench}.
We hope our findings guide future work on developing inclusive language technologies through high-quality synthetic data.

\section*{Limitations}

Our work comes with some limitations and open questions left for future work.
For example, our language set encompasses six languages.
Although we chose these languages carefully based on (1) whether they can be evaluated on publicly available LM benchmarks and (2) prior theoretical work on principled test language selection \citep{ploeger-etal-2026-principled},
validating our findings across a broader language sample remains important future work.
In addition, our \textit{Translate} data generation method assumes access to English prompts that can be meaningfully translated to target languages.
This approach inherits limitations from LM-based techniques, such as difficulty localizing culture-specific references and the risk of introducing translationese artifacts.

\section*{Ethics Statement}
Synthetic data generation risks amplifying biases present in teacher models.
If a teacher model underperforms on certain languages or exhibits cultural biases, these weaknesses propagate to student models trained on its outputs.
Our finding that teacher effectiveness correlates with CommonCrawl representation (\appref{sec:language}) suggests that already underrepresented languages may be further disadvantaged in synthetic data pipelines, potentially widening the performance gap between high- and low-resource languages.

\makeatletter
\ifacl@finalcopy
    \section*{Acknowledgments}

    LJVM and AK acknowledge the support of the UKRI Frontier Grant EP/Y031350/1 (EQUATE).
    This work was performed using joint resources provided by the Cambridge
    Service for Data Driven Discovery (CSD3) EP/T022159/1,
    Isambard AI National AI Research Resource (AIRR) ST/AIRR/I-A-I/1023, and
    the Microsoft Research Grant.
    LJVM would also like to thank Songbo Hu, Chen Cecilia Liu, Millicent Ochieng, and Felermino Ali for helpful and productive discussions on the project.
    Finally, the authors thank the reviewers and area chair of the May ARR cycle for valuable feedback that further improved this work.
\fi
\makeatother

% References

\bibliography{references}

\clearpage
\appendix

\onecolumn
\addtocontents{toc}{\protect\setcounter{tocdepth}{2}}
\section*{Appendix}
\renewcommand{\contentsname}{}
\tableofcontents
\twocolumn

\begin{table*}[t]
    \centering
    \footnotesize
    \renewcommand{\arraystretch}{0.80}
    \begin{tabularx}{\textwidth}{XXX}
        \toprule
        \textbf{Dataset} & \textbf{Language(s)} & \textbf{Generation Method / Description} \\
        \midrule
        % Add survey entries here
        % \begin{noindent}
        Bactrian-X \citep{li2023bactrian} & 52 languages - Arabic, Indonesian, Chinese, Malay, Tamil, Tagalog, etc. & \textit{Translate} - used Google Translate API to translate English instructions from Alpaca (52K) and Dolly (15K).                       \\
        MultiAlpaca \citep{wei2023polylmopensourcepolyglot} & 18 languages - English, Chinese, Russian, Spanish, German, French, etc. & \textit{Generate, Translate} - used a multilingual self-instruct \citep{wang-etal-2023-self-instruct} method from English prompt-response pairs to perform translation.  \\
        xP3-MT \citep{muennighoff-etal-2023-crosslingual} & 46 languages - Arabic, English, Spanish, Hindi, Chinese, Indonesian, etc. & \textit{Translate, Respond} - used Google Translate API to translate English prompt-response pairs from different sources, in addition to creating template-based prompts that an LM responds to. \\
        Cendol \citep{cahyawijaya-etal-2024-cendol} & 18 Indonesian languages - Sundanese, Javanese, Acehnese, Banjarese, Buginese, Gorontalo, etc. & \textit{Translate, Respond} - curated various prompts from past Indonesian NLP tasks, including translations of Dolly. \\
        Seed-Free Thai \citep{pengpun-etal-2024-seed} & Thai & \textit{Generate} - generated synthetic instruction data without seed examples by using Wikipedia contexts. Identifies fluency, diversity, and cultural context as key properties.\\
        Aya Dataset and Collection \citep{singh-etal-2024-aya} & 114 languages - Arabic, French, Hindi, Indonesian, Japanese, Spanish, Swahili, Turkish, Yoruba, Filipino, etc.& \textit{Translate, Respond} - involves a collection of translated prompts from English, and templated prompts. A sizeable portion of the collection includes native-speaker annotations. \\
        sPhinX \citep{ahuja-etal-2025-sphinx} & 51 languages - Afrikaans, Arabic, Bengali, Bulgarian, Burmese, Chinese, Croatian, Czech, etc. & \textit{Translate} - selectively translates essential portions of multilingual inputs in order to semantically preserve meaning. \\
        EuroBlocks \citep{martins2025eurollm,martins2024eurollmmultilinguallanguagemodels} & 31 languages - English, Chinese, Spanish, Italian, French, German, Portuguese, Dutch, Polish, etc. & \textit{Generate, Translate} - prompted Llama 3 or an earlier EuroLLM checkpoint with a document, target language, and category, then asked it to generate an instruction. Also involved translating prompt-response pairs. \\
        SEA-LION Dataset \citep{ng2025sea} & 11 languages - English, Chinese, Indonesian, Vietnamese, Malay, Thai, Burmese, Lao, Filipino, Khmer, and Tamil & \textit{Generate, Translate} - for the majority of the datasets, samples were first generated in English using Qwen 32B, and then translated into the target language using Gemma 2 27B. \\
        Urdu-Instruct Dataset \citep{shafique-etal-2025-alif} & Urdu & \textit{Generate} - uses a modified Self-Instruct from a pool of culturally relevant prompts.\\
        Pragyaan \citep{rachamalla-etal-2025-pragyaan} & 10 Indian languages - Gujarati, Kannada, Marathi, Bengali, Odia, Tamil, Malayalam, Telugu, Punjabi, and Hindi & \textit{Generate, Translate} - perform translation using an LM for a subset of data. Used Self-Instruct from a pool of native prompts for another subset of data. \\
        \bottomrule
        % \end{noindent}
    \end{tabularx}
    \caption{
        \textbf{Short survey of related work on synthetic data generation for multilingual LMs.}
        For each work, we provide a brief description of its data generation method.
        We find that most methods fall into one of the three categories described in \S\ref{sec:synthetic-data-generation}, i.e., Generate, Translate, or Respond, which we tested in our experiments.
    }
    \label{table:survey}
\end{table*}

\clearpage

% \section{Extended Related Work}

% \paragraph{Knowledge Distillation.}
% Synthetic data generation can also be considered as sequence-level knowledge distillation \citep{kim-rush-2016-sequence} or \textit{off-policy} distillation in reinforcement learning parlance \citep{lu2025onpolicydistillation}.
% In this setting, we collect outputs from some external source, i.e., the teacher model, that the student learns to imitate.
% Another setting is \textit{on-policy} distillation, where the student generates outputs that are scored or refined by the teacher, with optimization constrained by a KL-divergence penalty to a reference policy.
% Our work focuses on the off-policy setting because it is a more common practice in multilingual SFT \citep[\textit{inter alia}]{cahyawijaya-etal-2024-cendol,nguyen-etal-2024-democratizing,martins2025eurollm}.

\section{Multilingual Synthetic Data Generation}
\label{appendix:lit-review-multilingual-sdg}

We present an overview of prior works in \autoref{table:survey} that used synthetic data to train multilingual LMs.
In general, we find that most data generation methods fall into one of the three categories described in \S\ref{sec:synthetic-data-generation}, i.e., \textit{Generate}, \textit{Translate}, or \textit{Respond}, which we tested in our experiments.
Our survey suggests that our \textbf{choice of data generation methods is grounded in prior work} and covers the majority of approaches used in synthetic data generation.

\section{Seed Dataset Statistics}
\label{appendix:seed_dataset_statistics}

\autoref{table:seed_dataset} shows the statistics of the seed dataset used for synthetic data generation.

\begin{table*}[t]
    \centering
    \footnotesize
    \renewcommand{\arraystretch}{0.9}
    \setlength{\tabcolsep}{1.2pt}
    \resizebox{\textwidth}{!}{
        \begin{tabular}{@{}lrrrrrrr@{}}
            \toprule
                            & \multicolumn{7}{c}{\textbf{Language}}                                                                                                                                                 \\
            \cmidrule(lr){2-8}
            \textbf{Source} & \textbf{English (en)}                 & \textbf{Arabic (ar)} & \textbf{Czech (cs)} & \textbf{German (de)} & \textbf{Spanish (es)} & \textbf{Indonesian (id)} & \textbf{Japanese (ja)} \\
            \midrule
            %\begin{noindent}
        Aya Dataset  & - & - & 5,000 & 241 & 3,854 & 2,786 & 6,259 \\
        T\"ulu 3 SFT   & 10,000 & - & - & - & - & - & - \\
        WildChat 4.8M  & 10,000 & 4,660 & 1,266 & 5,908 & 5,900 & 7,983 & 602 \\
        CIDAR & - & 6,000 & - & - & - & - & - \\
        Cendol v2 & - & - & - & - & - & 3,000 & - \\
        OpenAssistant 2 & - & 23 & 4 & 2,328 & 8,785 & 3 & 306 \\
        EuroBlocks SFT & - & - & 3,813 & 12,551 & 15,641 & - & 2,893 \\
        GSM8K (train) & 7,473 & - & - & - & - & - & - \\
        Helpsteer3 (chosen) & - & - & - & 462 & 778 & 156 & 534 \\
        Magpie Pro Filtered & 10,000 & - & - & - & - & - & - \\
        \midrule
        \textbf{Total per language} & 37,473 & 10,683 & 10,083 & 21,490 & 34,958 & 13,928 & 10,594 \\
        % \textbf{Total examples} & \multicolumn{7}{r}{\textbf{129,929}} \\
        %\end{noindent}
            \bottomrule
        \end{tabular}
    }
    \caption{
        \textbf{Seed dataset statistics.}
        In order to bootstrap our synthetic data generation methods, we use a seed dataset composed of various multilingual instruction-following datasets.
        We include English samples in order to simulate data generation pipelines where English is translated into a target language.
        We collect a total of \textbf{139,209 seed examples} across 7 languages (including English).
    }
    \label{table:seed_dataset}
\end{table*}

\section{Release of Data and Software Artifacts}
\label{appendix:polyglot_collection}
In order to facilitate future research on multilingual synthetic data generation, we introduce the \textsc{Polyglot} collection, a collection of synthetic datasets produced by the best teacher model across all target languages, together with the student models finetuned on them.
The \textsc{Polyglot} collection includes:

\begin{itemize}[complist]
    \item PolyglotTeachers-SFT-Synth-Data: Synthetic datasets for each target language generated by each teacher model using all three data generation methods (\S\ref{sec:synthetic-data-generation}).
    \item PolyglotTeachers-Multilingual-Instruct: A set of 7B student models finetuned on each synthetic dataset from the \basemodel{} base model using the Gemma 3 27B (highest-scoring model) teacher.
\end{itemize}

Because synthetic data generation can be expensive \citep{lambert2026resourceefficientllmsendtoendenergy}, especially when accounting for teacher inference costs, we hope that the \textsc{Polyglot} Collection can help amortize the cost of obtaining quality multilingual data without re-running these systematic experiments.
\makeatletter
\ifacl@finalcopy
    We publicly release these artifacts on HuggingFace.\footnote{\href{https://huggingface.co/collections/ljvmiranda921/polyglot-teachers}{\huggingface{}: \texttt{ljvmiranda921/polyglot-teachers}}}
\else
    % use this
    We will publicly release the \textsc{Polyglot} Collection after the review period.
\fi
\makeatother

\section{Teacher Model and Target Language Details}
\label{appendix:teacher_target_details}

In this section, we provide additional details about the teacher models and target languages used in our experiments.
\autoref{table:teacher_model_details} summarizes the key characteristics of each teacher model.
On the other hand, \autoref{table:language_details} provides information about the target languages, including language family, number of speakers, and resource availability.

\begin{table*}[t]
    \centering
    %\footnotesize
    \begin{tabular}{@{}lllrl@{}}
        \toprule
        \textbf{Model Name}                                 & \textbf{Provider} & \textbf{Size (B)} & \textbf{\# Langs} & \textbf{License} \\
        \midrule
        GPT-4o mini \citep{hurst2024gpt}                    & OpenAI            & --                & 50+               & Proprietary      \\
        % Claude Sonnet 4.5 \citep{claude3modelcard2024}      & Anthropic         & --                & 50+               & Proprietary      \\
        % Claude Haiku 4.5 \citep{claude3modelcard2024}       & Anthropic         & --                & 50+               & Proprietary      \\
        Llama 3.1 70B Instruct \citep{grattafiori2024llama} & Meta              & 70                & 8                 & Llama 3.1        \\
        Llama 3.1 8B Instruct \citep{grattafiori2024llama}  & Meta              & 8                 & 8                 & Llama 3.1        \\
        Command A \citep{cohere2025command}                 & Cohere            & 104               & 23                & CC-BY-NC-4.0     \\
        Aya Expanse 32B \citep{dang2024aya}                 & Cohere            & 32                & 23                & CC-BY-NC-4.0     \\
        Gemma 3 27B Instruct \citep{team2025gemma}          & Google            & 27                & 100+              & Gemma            \\
        Gemma 3 12B Instruct \citep{team2025gemma}          & Google            & 12                & 100+              & Gemma            \\
        Gemma 3 4B Instruct \citep{team2025gemma}           & Google            & 4                 & 100+              & Gemma            \\
        IBM Granite 4.0 \citep{ibm-granite4.0-2025}         & IBM               & 3                 & 116               & Apache 2.0       \\
        IBM Granite Micro \citep{ibm-granite4.0-2025}       & IBM               & 0.4               & 116               & Apache 2.0       \\
        \bottomrule
    \end{tabular}
    \caption{
        \textbf{Teacher model details.}
        We evaluate \nummodels{} teacher models across different providers, sizes, multilingual capabilities, and licensing terms.
        Size is reported in billions of parameters (B) where available.
        \# Langs indicates the number of languages the model was trained on or evaluated for.
    }
    \label{table:teacher_model_details}
\end{table*}

\begin{table*}[t]
    \centering
    \begin{tabular}{@{}llllr@{}}
        \toprule
        \textbf{Language} & \textbf{Family} & \textbf{Script} & \textbf{Resource Availability} & \textbf{\% in CC} \\
        \midrule
        Arabic            & Afro-Asiatic    & Arabic          & 5 (High)                       & 0.65\%            \\
        Czech             & Indo-European   & Latin           & 4 (Medium-High)                & 0.99\%            \\
        German            & Indo-European   & Latin           & 5 (High)                       & 6.01\%            \\
        Spanish           & Indo-European   & Latin           & 5 (High)                       & 4.37\%            \\
        Indonesian        & Austronesian    & Latin           & 3 (Medium)                     & 0.95\%            \\
        Japanese          & Japonic         & Japanese        & 5 (High)                       & 5.20\%            \\
        \bottomrule
    \end{tabular}
    \caption{
        \textbf{Target language details.}
        We evaluate teacher models across six typologically diverse languages spanning different language families and scripts.
        Resource availability is based on the classification from \citet{joshi-etal-2020-state}, ranging from 0 (lowest) to 5 (highest).
        CommonCrawl percentages \citep{raffel2023exploringlimitstransferlearning} indicate the proportion of web text available for each language.
    }
    \label{table:language_details}
\end{table*}

\section{Experimental Details}

\subsection{Supervised Finetuning}
\label{appendix:finetuning_details}

\autoref{table:finetuning_hyperparameters} summarizes the hyperparameters used for finetuning student models.
We train models with the Unsloth framework \citep{unsloth} on a cluster of Grace Hopper GH200 Superchips.
Full finetuning (7B) takes around 1.5 hours (wall clock) for 2 epochs on 2 nodes.

\begin{table}[H]
    \centering
    \resizebox{\columnwidth}{!}{%
        \begin{tabular}{lrlr}
            \toprule
            \textbf{Hyperparameter} & \textbf{Value} & \textbf{Hyperparameter} & \textbf{Value} \\
            \midrule
            Learning rate           & 5e-5           & Batch size              & 32             \\
            Epochs                  & 2              & Grad. Acum. Steps       & 4              \\
            Max seq. length         & 16,384         & Weight decay            & 0.001          \\
            Optimizer               & AdamW          & Scheduler               & Linear         \\
            \bottomrule
        \end{tabular}%
    }
    \caption{Hyperparameters for finetuning a 7B student model from \basemodel{}.}
    \label{table:finetuning_hyperparameters}
\end{table}

\subsection{Model Evaluation}
\label{appendix:evaluation_details}

We use the Lighteval framework \citep[\texttt{v0.13.1dev0},][]{lighteval} for evaluation.
\autoref{table:evaluation_details} summarizes the benchmarks used for evaluating student models.
We use Global-MMLU \textit{Lite} instead of Global-MMLU because the former contains actual native speaker annotations that localized the benchmark into different cultural contexts.

\begin{table}[H]
    \centering
    \resizebox{\columnwidth}{!}{%
        \begin{tabular}{lllr}
            \toprule
            \textbf{Benchmark} & \textbf{Formulation} & \textbf{Metric} & \textbf{N-shots} \\
            \midrule
            Global-MMLU Lite   & MCF                  & Accuracy        & 0                \\
            M-RewardBench      & MCF                  & Weighted Acc.   & 0                \\
            M-GSM              & Generative           & Exact-Match     & 5                \\
            \bottomrule
        \end{tabular}%
    }
    \caption{Evaluation settings for each benchmark (MCF: Multiple-Choice Formulation).}
    \label{table:evaluation_details}
\end{table}

For Global-MMLU Lite and M-RewardBench, we use the Multiple-Choice Formulation (MCF) with character normalization.
In addition, we also follow the corpus-level metric in M-RewardBench which uses a weighted accuracy for each data subset and category \citep{gureja-etal-2025-rewardbench}.
For M-GSM, we show 5 few-shot examples from the training set in order for the model to properly generate the answer.
We run all evaluation experiments for three trials with different random seeds and report the average and standard error.

\section[Full Results for Intrinsic and Extrinsic Metrics]{Full Results for Intr. and Extr. Metrics}

\autoref{table:full_intrinsic_results} shows all the data quality metrics for each teacher model across all languages.
\autoref{table:full_student_model_results} shows the full results of student models finetuned on synthetic datasets generated by each teacher model across all target languages.

\begin{table*}[t]
    \centering
    % \begin{noindent}
    \footnotesize
    \begin{tabular}{@{}lcccccccccccc@{}}
        \toprule
        & \multicolumn{4}{c}{\textbf{Arabic (ar)}} & \multicolumn{4}{c}{\textbf{Czech (cs)}} & \multicolumn{4}{c}{\textbf{German (de)}} \\
        \cmidrule(lr){2-5} \cmidrule(lr){6-9} \cmidrule(lr){10-13}
        \textbf{Model} & $d_x$ & $d_y$ & PPL & R & $d_x$ & $d_y$ & PPL & R & $d_x$ & $d_y$ & PPL & R \\
        \midrule
        GPT-4o mini & 0.704 & 0.869 & 8.40 & 3.516 & 0.643 & 0.862 & 3.18 & 3.716 & 0.732 & 0.889 & 3.65 & 3.810 \\
        Llama 3.1 70B Inst. & 0.701 & 0.875 & 7.00 & 2.719 & 0.654 & 0.889 & 3.18 & 3.327 & 0.707 & 0.892 & 3.22 & 3.396 \\
        Llama 3.1 8B Inst. & 0.708 & 0.779 & 6.2e4 & 1.731 & 0.673 & 0.799 & 2.7e4 & 1.908 & 0.738 & 0.873 & 3.6e3 & 2.513 \\
        Command A & 0.690 & 0.846 & 5.41 & 3.996 & 0.647 & 0.865 & 3.24 & 4.184 & 0.730 & 0.889 & 3.59 & 4.235 \\
        Aya Expanse 32B & 0.693 & 0.888 & 4.34 & 3.964 & 0.650 & 0.884 & 3.15 & 4.133 & 0.700 & 0.902 & 3.44 & 4.140 \\
        Gemma 3 27B Inst. & 0.717 & 0.890 & 4.40 & 3.932 & 0.675 & 0.885 & 3.77 & 4.342 & 0.731 & 0.898 & 3.96 & 4.260 \\
        Gemma 3 12B Inst. & 0.721 & 0.864 & 4.43 & 3.774 & 0.676 & 0.882 & 3.88 & 4.266 & 0.751 & 0.899 & 4.06 & 4.203 \\
        Gemma 3 4B Inst. & 0.728 & 0.869 & 5.52 & 3.470 & 0.682 & 0.883 & 3.87 & 4.127 & 0.744 & 0.898 & 3.96 & 4.103 \\
        IBM Granite 4.0 & 0.704 & 0.829 & 1.9e4 & 2.463 & 0.665 & 0.862 & 5.29 & 3.158 & 0.717 & 0.885 & 24.61 & 3.365 \\
        IBM Granite Micro & 0.741 & 0.863 & 12.45 & 3.033 & 0.713 & 0.874 & 4.61 & 3.568 & 0.726 & 0.892 & 4.59 & 3.704 \\
        \bottomrule
    \end{tabular}

    \vspace{1em}

    \begin{tabular}{@{}lcccccccccccc@{}}
        \toprule
        & \multicolumn{4}{c}{\textbf{Spanish (es)}} & \multicolumn{4}{c}{\textbf{Indonesian (id)}} & \multicolumn{4}{c}{\textbf{Japanese (ja)}} \\
        \cmidrule(lr){2-5} \cmidrule(lr){6-9} \cmidrule(lr){10-13}
        \textbf{Model} & $d_x$ & $d_y$ & PPL & R & $d_x$ & $d_y$ & PPL & R & $d_x$ & $d_y$ & PPL & R \\
        \midrule
        GPT-4o mini & 0.729 & 0.887 & 3.78 & 3.883 & 0.728 & 0.854 & 5.50 & 3.656 & 0.736 & 0.880 & 5.81 & 3.639 \\
        Llama 3.1 70B Inst. & 0.728 & 0.892 & 3.15 & 3.434 & 0.727 & 0.874 & 4.85 & 3.293 & 0.756 & 0.799 & 4.52 & 2.459 \\
        Llama 3.1 8B Inst. & 0.744 & 0.898 & 503.0 & 2.860 & 0.738 & 0.863 & 1.1e3 & 2.599 & 0.759 & 0.796 & 5.4e4 & 1.806 \\
        Command A & 0.733 & 0.884 & 3.77 & 4.336 & 0.747 & 0.857 & 4.94 & 3.899 & 0.739 & 0.881 & 4.92 & 4.174 \\
        Aya Expanse 32B & 0.724 & 0.893 & 3.67 & 4.181 & 0.726 & 0.879 & 4.43 & 4.017 & 0.743 & 0.883 & 5.96 & 3.821 \\
        Gemma 3 27B Inst. & 0.768 & 0.903 & 4.30 & 4.266 & 0.740 & 0.854 & 5.49 & 4.057 & 0.765 & 0.875 & 5.90 & 3.956 \\
        Gemma 3 12B Inst. & 0.763 & 0.895 & 4.14 & 4.193 & 0.762 & 0.851 & 5.84 & 3.958 & 0.756 & 0.885 & 5.78 & 4.017 \\
        Gemma 3 4B Inst. & 0.754 & 0.887 & 4.52 & 4.021 & 0.760 & 0.851 & 6.46 & 3.657 & 0.794 & 0.875 & 6.45 & 3.656 \\
        IBM Granite 4.0 & 0.743 & 0.882 & 5.22 & 3.309 & 0.729 & 0.833 & 16.80 & 2.437 & 0.761 & 0.849 & 9.79 & 2.889 \\
        IBM Granite Micro & 0.729 & 0.887 & 4.58 & 3.779 & 0.760 & 0.860 & 11.92 & 3.113 & 0.764 & 0.877 & 7.22 & 3.295 \\
        \bottomrule
    \end{tabular}
    %\end{noindent}
    \caption{
        \textbf{Full intrinsic evaluation results across all languages.}
        Data quality metrics include the diversity of prompts and responses ($d_x$ and $d_y$), average perplexity of the base model on the response (PPL), and average reward score based on a multilingual LLM judge (R).
    }
    \label{table:full_intrinsic_results}
\end{table*}
\definecolor{cambridgeblue}{HTML}{254EFF}
\definecolor{warmcrest}{HTML}{C96A2E}

\begin{table*}[t]
    \centering
    % \begin{noindent}
    \footnotesize
    \begin{tabular}{@{}lcccccc@{}}
        \toprule
        \textbf{Model} & \textbf{Arabic (ar)} & \textbf{Czech (cs)} & \textbf{German (de)} & \textbf{Spanish (es)} & \textbf{Indonesian (id)} & \textbf{Japanese (ja)} \\
        \midrule
        GPT-4o mini & \cellcolor{warmcrest!34}$-$2.086 & \cellcolor{cambridgeblue!8}0.538 & \cellcolor{cambridgeblue!50}3.098 & \cellcolor{cambridgeblue!23}1.395 & \cellcolor{cambridgeblue!33}2.025 & \cellcolor{cambridgeblue!1}0.099 \\
        Llama 3.1 70B Inst. & \cellcolor{warmcrest!25}$-$1.528 & \cellcolor{cambridgeblue!8}0.538 & \cellcolor{cambridgeblue!37}2.265 & \cellcolor{cambridgeblue!17}1.075 & \cellcolor{cambridgeblue!5}0.329 & \cellcolor{white}0.013 \\
        Llama 3.1 8B Inst. & \cellcolor{warmcrest!14}$-$0.841 & \cellcolor{cambridgeblue!8}0.525 & \cellcolor{cambridgeblue!43}2.623 & \cellcolor{cambridgeblue!9}0.595 & \cellcolor{cambridgeblue!23}1.425 & \cellcolor{cambridgeblue!3}0.236 \\
        Command A & \cellcolor{warmcrest!41}$-$2.476 & \cellcolor{cambridgeblue!8}0.505 & \cellcolor{cambridgeblue!45}2.759 & \cellcolor{cambridgeblue!26}1.613 & \cellcolor{cambridgeblue!31}1.863 & \cellcolor{cambridgeblue!14}0.841 \\
        Aya Expanse 32B & \cellcolor{warmcrest!4}$-$0.293 & \cellcolor{cambridgeblue!8}0.538 & \cellcolor{cambridgeblue!41}2.491 & \cellcolor{cambridgeblue!28}1.701 & \cellcolor{cambridgeblue!32}1.943 & \cellcolor{cambridgeblue!3}0.221 \\
        Gemma 3 27B Inst. & \cellcolor{warmcrest!1}$-$0.074 & \cellcolor{cambridgeblue!9}0.552 & \cellcolor{cambridgeblue!43}2.635 & \cellcolor{cambridgeblue!28}1.724 & \cellcolor{cambridgeblue!3}0.198 & \cellcolor{cambridgeblue!11}0.677 \\
        Gemma 3 12B Inst. & \cellcolor{warmcrest!16}$-$1.015 & \cellcolor{cambridgeblue!8}0.538 & \cellcolor{cambridgeblue!44}2.700 & \cellcolor{cambridgeblue!26}1.592 & \cellcolor{white}$-$0.017 & \cellcolor{cambridgeblue!8}0.524 \\
        Gemma 3 4B Inst. & \cellcolor{warmcrest!17}$-$1.033 & \cellcolor{cambridgeblue!8}0.538 & \cellcolor{cambridgeblue!42}2.568 & \cellcolor{cambridgeblue!20}1.209 & \cellcolor{warmcrest!6}$-$0.388 & \cellcolor{cambridgeblue!5}0.349 \\
        IBM Granite 4.0 & \cellcolor{cambridgeblue!26}1.565 & \cellcolor{cambridgeblue!8}0.538 & \cellcolor{cambridgeblue!34}2.061 & \cellcolor{cambridgeblue!20}1.235 & \cellcolor{cambridgeblue!10}0.614 & \cellcolor{cambridgeblue!13}0.802 \\
        IBM Granite Micro & \cellcolor{warmcrest!7}$-$0.421 & \cellcolor{cambridgeblue!8}0.538 & \cellcolor{cambridgeblue!30}1.842 & \cellcolor{cambridgeblue!20}1.203 & \cellcolor{warmcrest!10}$-$0.659 & \cellcolor{cambridgeblue!3}0.210 \\
        \bottomrule
    \end{tabular}
    % \end{noindent}
    \caption{
        \textbf{Average performance gap recovered (PGR) of a student model across various multilingual benchmarks.}
        Our multilingual evaluation suite includes Global-MMLU Lite \citep{singh-etal-2025-global}, M-RewardBench \citep{gureja-etal-2025-rewardbench}, and M-GSM \citep{shi2023language}.
        The PGR computation is based on \citet{kim-etal-2025-evaluating} and detailed in \S\ref{sec:extrinsic-metrics} (\autoref{eq:extrinsic_score}) where $\refstudent=\text{OLMo 3 7B Instruct SFT}$ and $\basestudent=\text{OLMo 3 7B}$.
    }
    \label{table:full_student_model_results}
\end{table*}

\definecolor{cambridgeblue}{HTML}{254EFF}
\definecolor{warmcrest}{HTML}{C96A2E}

\begin{table*}[t]
    \centering
    \resizebox{\textwidth}{!}{%
        \begin{tabular}{@{}lcccccc@{}}
            \toprule
            \textbf{Model}      & \textbf{Arabic (ar)}                               & \textbf{Czech (cs)}                                   & \textbf{German (de)}                                   & \textbf{Spanish (es)}                                  & \textbf{Indonesian (id)}                               & \textbf{Japanese (ja)}                                 \\
            \midrule
            Gemma 3 27B Inst.   & \cellcolor{cambridgeblue!2}\textbf{0.145 (0.0121)} & \cellcolor{cambridgeblue!6}\textbf{0.360 (0.0004)}    & \cellcolor{cambridgeblue!27}1.655 (0.0141)             & \cellcolor{cambridgeblue!22}\textbf{1.358 (0.0141)}    & \cellcolor{cambridgeblue!3}0.214 (0.0167)              & \cellcolor{cambridgeblue!10}\underline{0.626 (0.0124)} \\
            Aya Expanse 32B     & \cellcolor{white}\underline{$-$0.058 (0.0116)}       & \cellcolor{cambridgeblue!3}0.222 (0.0004)             & \cellcolor{cambridgeblue!24}1.468 (0.0134)             & \cellcolor{cambridgeblue!18}1.129 (0.0123)             & \cellcolor{cambridgeblue!19}\textbf{1.153 (0.0124)}    & \cellcolor{cambridgeblue!5}0.320 (0.0111)              \\
            Gemma 3 12B Inst.   & \cellcolor{warmcrest!7}$-$0.464 (0.0119)             & \cellcolor{cambridgeblue!5}\underline{0.327 (0.0004)} & \cellcolor{cambridgeblue!29}\underline{1.756 (0.0137)} & \cellcolor{cambridgeblue!20}\underline{1.228 (0.0140)} & \cellcolor{cambridgeblue!2}0.151 (0.0126)              & \cellcolor{cambridgeblue!9}0.573 (0.0142)              \\
            Command A           & \cellcolor{warmcrest!22}$-$1.360 (0.0112)            & \cellcolor{cambridgeblue!1}0.114 (0.0004)             & \cellcolor{cambridgeblue!27}1.673 (0.0139)             & \cellcolor{cambridgeblue!18}1.102 (0.0145)             & \cellcolor{cambridgeblue!17}\underline{1.063 (0.0125)} & \cellcolor{cambridgeblue!11}\textbf{0.683 (0.0122)}    \\
            Gemma 3 4B Inst.    & \cellcolor{warmcrest!8}$-$0.488 (0.0119)             & \cellcolor{cambridgeblue!5}0.330 (0.0004)             & \cellcolor{cambridgeblue!27}1.644 (0.0137)             & \cellcolor{cambridgeblue!15}0.929 (0.0140)             & \cellcolor{warmcrest!1}$-$0.105 (0.0126)                 & \cellcolor{cambridgeblue!8}0.504 (0.0113)              \\
            GPT-4o mini         & \cellcolor{warmcrest!18}$-$1.117 (0.0117)            & \cellcolor{white}0.015 (0.0004)                       & \cellcolor{cambridgeblue!29}\textbf{1.766 (0.0136)}    & \cellcolor{cambridgeblue!15}0.908 (0.0149)             & \cellcolor{cambridgeblue!16}1.003 (0.0125)             & \cellcolor{cambridgeblue!3}0.189 (0.0117)              \\
            IBM Granite 4.0     & \cellcolor{warmcrest!1}$-$0.072 (0.0123)             & \cellcolor{white}$-$0.031 (0.0004)                      & \cellcolor{cambridgeblue!16}1.000 (0.0135)             & \cellcolor{cambridgeblue!12}0.734 (0.0151)             & \cellcolor{warmcrest!1}$-$0.079 (0.0125)                 & \cellcolor{cambridgeblue!5}0.321 (0.0108)              \\
            IBM Granite Micro   & \cellcolor{warmcrest!4}$-$0.282 (0.0121)             & \cellcolor{cambridgeblue!4}0.290 (0.0004)             & \cellcolor{cambridgeblue!18}1.102 (0.0139)             & \cellcolor{cambridgeblue!13}0.783 (0.0133)             & \cellcolor{warmcrest!5}$-$0.329 (0.0126)                 & \cellcolor{cambridgeblue!4}0.264 (0.0121)              \\
            Llama 3.1 70B Inst. & \cellcolor{warmcrest!16}$-$0.964 (0.0117)            & \cellcolor{cambridgeblue!1}0.109 (0.0004)             & \cellcolor{cambridgeblue!19}1.195 (0.0146)             & \cellcolor{cambridgeblue!11}0.688 (0.0146)             & \cellcolor{cambridgeblue!3}0.182 (0.0126)              & \cellcolor{warmcrest!6}$-$0.373 (0.0116)                 \\
            Llama 3.1 8B Inst.  & \cellcolor{warmcrest!28}$-$1.693 (0.0120)            & \cellcolor{warmcrest!16}$-$0.974 (0.0004)               & \cellcolor{cambridgeblue!14}0.891 (0.0148)             & \cellcolor{cambridgeblue!3}0.182 (0.0164)              & \cellcolor{cambridgeblue!5}0.322 (0.0124)              & \cellcolor{warmcrest!14}$-$0.863 (0.0129)                \\
            \bottomrule
        \end{tabular}
    }%
    \caption{
        \textbf{Detailed results from \autoref{table:sota_evals} with standard errors.}
        We compute \shortmethodname{} thrice with different synthetically generated data (each trial uses a different data mix based on a random seed).
        We report the mean and standard error for each teacher model across all target languages.
        For each language, we highlight the best model in \textbf{bold} and the second-best model with an \underline{underline}.
    }
    \label{table:sota_evals_appendix}
\end{table*}

\paragraph{Percentage Increase Tables}
We provide additional tables from the main experiments in \S\ref{sec:experiments} and \S\ref{sec:analysis}.
\autoref{tab:pct_increase_data_generation_method} shows the percentage increase in \shortmethodname{} when using the best data generation method for each teacher-language pair compared to an equal mix baseline (see \S\ref{sec:effect_generation_method}).

\begin{table*}[t]
    \centering
    {
        %\footnotesize
        \setlength{\tabcolsep}{4pt}
        \begin{tabular}{@{}llrrr@{}}
            \toprule
                              &                      & \multicolumn{3}{c}{\textbf{Teacher Model} ($\teacher$)}                                                     \\
            \cmidrule(lr){3-5}
            \textbf{Language} & \textbf{Best Method} & \textbf{Gemma 3 27B}                                      & \textbf{Aya Expanse 32B} & \textbf{Llama 3.1 70B} \\
            \midrule
            Arabic (ar)       & Respond              & +453.1\%                                                  & +355.2\%                 & +77.7\%                \\
            German (de)       & Generate             & +29.3\%                                                   & +0.3\%                   & +16.4\%                \\
            Indonesian (id)   & Translate            & +458.9\%                                                  & +39.3\%                  & $-$14.8\%              \\
            \bottomrule
        \end{tabular}
    }
    \caption{
        \textbf{Percentage increase in \shortmethodname{} for best data generation method.}
        Percentage increase when using the best-performing data generation method compared to an equal mix baseline of all three methods (Generate, Translate, Respond).
        For less-resourced languages (Arabic and Indonesian), using Translate or Respond methods yields substantial improvements for most teachers, though gains are teacher-dependent.
    }
    \label{tab:pct_increase_data_generation_method}
\end{table*}

\section{Additional Experiments and Ablations}
\label{appendix:additional_experiments}

In this section, we ablate several aspects of our evaluation protocol that may affect a teacher model's \shortmethodname{}.

\subsection{Effect of Data Scale on Student Model Performance}
\label{appendix:data_scale}

One component of \shortmethodname{} is the extrinsic student performance metric (\S\ref{sec:extrinsic-metrics}) as measured by PGR.
Scaling laws suggest that this performance improves with more data \citep{kaplan2020scalinglawsneurallanguage}.
Then, it is possible to inflate \shortmethodname{} by simply using more synthetic data.
In order to control for this variable, we conduct an experiment to determine how much synthetic data is needed to reliably compute \shortmethodname{}.

\paragraph{Setup}
We finetune an \basemodel{} base model on $n$ SFT instances where $n \in \{\text{1k}, \text{5k}, \text{10k}, \text{25k}, \text{50k}\}$.
To reduce computational costs, we perform this experiment only on a single teacher model (Gemma 3 27B Instruct) and three target languages that represent diverse scripts and resource availability: Arabic, German, and Indonesian.
Similar to the main experiments, we represent each data generation method equally when creating the SFT datasets.
Then, we recompute the intrinsic metrics and finetune student models and measure their performance across three benchmarks (\S\ref{sec:extrinsic-metrics}).

\begin{figure}[t]
    \centering
    \includegraphics[width=0.95\linewidth, trim={0cm 0cm 0cm 0cm}]{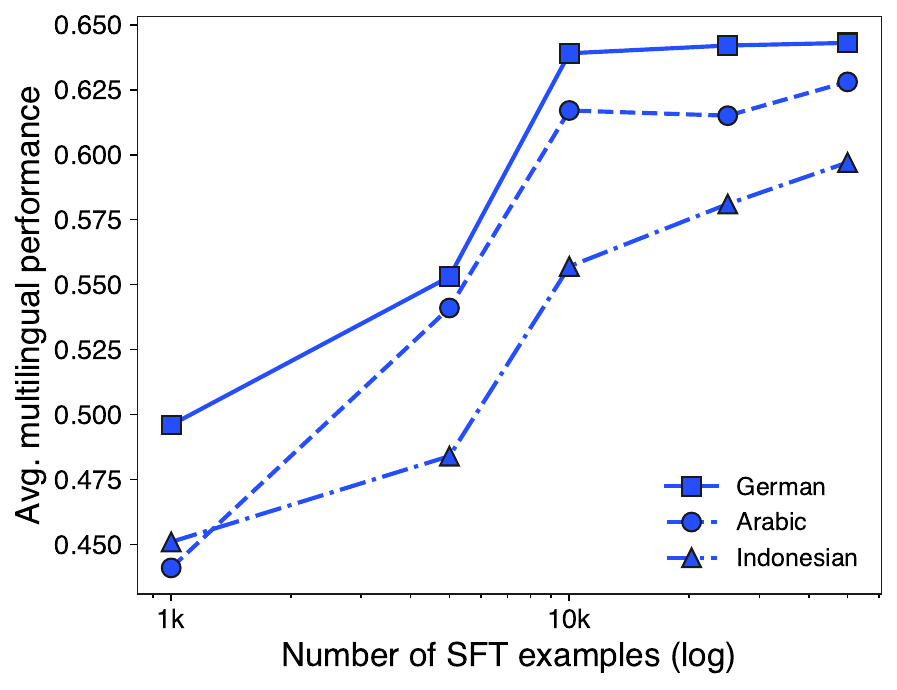}
    \caption{
        \textbf{Effect of synthetic data scale on student model performance.}
        Performance improves with more synthetic data, but gains diminish beyond 10k examples.
    }
    \label{fig:data_scale_effect}
\end{figure}

\paragraph{Results}
\autoref{fig:data_scale_effect} shows the average student model performance as a function of the number of SFT instances.
We observe that student performance improves with more synthetic data, but gains diminish beyond 10k examples.
This finding suggests that \textbf{using 10k synthetic examples per language is sufficient to reliably compute \shortmethodname{} without inflating the metric} by increasing the number of samples.
In our main experiments, we use 10.5k synthetic examples per language (3.5k per data generation method) when computing \shortmethodname{}.
%This finding also aligns with prior work \citep[\textit{Superficial Alignment Hypothesis, LIMA}][]{zhou2023lima}, where a small number of high-quality examples is enough to tune a pretrained language model.
Specifically, we show that 10k synthetic examples from a strong teacher are sufficient to finetune a student model to achieve reasonable performance across multiple benchmarks.

\subsection{Generalization Across Model Size}
\label{appendix:model_scale}

\paragraph{Setup}
In order to test whether \shortmethodname{} generalizes beyond 7B parameter size models,
we use an OLMo 3 32B base model ($\basestudent$) and recompute the intrinsic and extrinsic metrics to obtain the \shortmethodname{}.
To reduce computational costs, we train student models across three teachers (Gemma 3 27B Instruct, Aya Expanse 32B, Llama 3.1 70B Instruct) and all \numlanguages{} target languages.

\definecolor{cambridgeblue}{HTML}{254EFF}
\definecolor{warmcrest}{HTML}{C96A2E}

\begin{table*}[t]
    \centering
    % \begin{noindent}
    %\renewcommand{\arraystretch}{0.9}
    \resizebox{\textwidth}{!}{%
    \begin{tabular}{@{}lw{c}{1.2cm}ccccccc@{}}
        \toprule
        \textbf{Teacher Model} & \textbf{Average} & \textbf{Arabic (ar)} & \textbf{Czech (cs)} & \textbf{German (de)} & \textbf{Spanish (es)} & \textbf{Indonesian (id)} & \textbf{Japanese (ja)} \\
        \midrule
        Gemma 3 27B Inst. & \cellcolor{cambridgeblue!16}0.805 & \cellcolor{warmcrest!4}$-$0.239 & \cellcolor{cambridgeblue!4}0.222 & \cellcolor{cambridgeblue!40}2.389 & \cellcolor{cambridgeblue!32}1.855 & \cellcolor{cambridgeblue!4}0.239 & \cellcolor{cambridgeblue!7}0.366 \\
        Aya Expanse 32B & \cellcolor{cambridgeblue!4}0.227 & \cellcolor{warmcrest!17}$-$0.872 & \cellcolor{white}$-$0.038 & \cellcolor{cambridgeblue!34}1.979 & \cellcolor{cambridgeblue!25}1.353 & \cellcolor{warmcrest!4}$-$0.249 & \cellcolor{warmcrest!16}$-$0.809 \\
        Llama 3.1 70B Inst. & \cellcolor{warmcrest!5}$-$0.267 & \cellcolor{warmcrest!30}$-$1.688 & \cellcolor{warmcrest!16}$-$0.807 & \cellcolor{cambridgeblue!16}0.838 & \cellcolor{cambridgeblue!26}1.407 & \cellcolor{warmcrest!26}$-$1.441 & \cellcolor{cambridgeblue!1}0.089 \\
        \bottomrule
    \end{tabular}
    }%
    % \end{noindent}
    \caption{
        \textbf{\shortmethodname{} of three teacher models ($\basestudent=\text{OLMo 3 32B}$).}
        We show that our findings generalize up to the 32B parameter range on the three teacher models we tested:
        (1) Gemma 3 27B maintains its position as the most effective teacher, and the (2) language-dependent effects are still apparent with German having the highest \shortmethodname{}s across most teachers.
    }
    \label{table:sota_evals_32b}
\end{table*}

\paragraph{Results}
\autoref{table:sota_evals_32b} shows the \shortmethodname{} scores for three teacher models when using OLMo 3 32B as the student model.
We find that \textbf{Gemma 3 27B Instruct remains the highest-scoring teacher in this comparison}, achieving the highest average \shortmethodname{} of 0.805 across all languages.
This result is consistent with our findings using the 7B student model (\S\ref{sec:experiments}), demonstrating that the superior data quality generated by Gemma 3 27B generalizes across model scales.
Aya Expanse 32B achieves a positive average \shortmethodname{} of 0.227, while Llama 3.1 70B Instruct shows a negative average of $-$0.267.

Furthermore, the \textbf{language-dependent effects observed in the 7B experiments remain consistent at the 32B scale}.
German continues to show the highest \shortmethodname{} values across all three teachers (2.389 for Gemma, 1.979 for Aya, 0.838 for Llama), suggesting that certain languages benefit more from synthetic data regardless of student model size.
Similarly, Spanish exhibits strong performance across all teachers, with \shortmethodname{} values ranging from 1.353 to 1.855.
In contrast, Arabic shows the most variable results, with Gemma achieving slightly negative scores ($-$0.239) while Aya and Llama show substantially lower performance ($-$0.872 and $-$1.688, respectively).
Overall, these findings demonstrate that \shortmethodname{} and teacher model rankings generalize to the 32B parameter range.

\subsection{Effect of Translation Method (Prompting an LM vs. Translation Model)}
\label{appendix:translation_method}

An alternative to using an LM for translating texts from English to a target language is via a translation model such as NLLB \citep{costa2022no}.
%Compared to prompting an LM to translate, these translation models are task-specific and made of an encoder-decoder architecture.
In this section, we examine the effect of the translation method on the \shortmethodname{} of teacher models.

\paragraph{Setup}
First, we filter and sample 10k English prompt-response pairs from the T\"ulu 3 SFT dataset.\footnote{T\"ulu 3 also contains non-English data. We perform English-language filtering using fastText \citep{joulin2016fasttextzipcompressingtextclassification,joulin-etal-2017-bag} and the \texttt{staticvectors} library.}
Then, using the NLLB model (\texttt{nllb-200-distilled-600M}), we perform two translation methods: (1) \textit{NLLB-Translate-then-Respond:} translate the prompts to each target language and prompt Gemma 3 27B Instruct to generate a response, and
(2) \textit{NLLB-Translate-Both}: translate both the prompts and responses from English to the target language.
We choose the 600M version due to its computational efficiency and popularity among practitioners, as measured by HuggingFace downloads and community likes.

We compare these methods against our original \textit{Translate} method, i.e., prompting Gemma 3 27B Instruct to directly translate the prompt and generate the response in the target language (\textit{LM-Translate}).
Then, we compute the intrinsic data quality metrics and finetune \basemodel{} student models on each synthetic dataset to compute \shortmethodname{}.

\begin{figure}[t]
    \centering
    \includegraphics[height=10cm, width=0.43\textwidth, trim={0cm 0cm 0cm 0cm}]{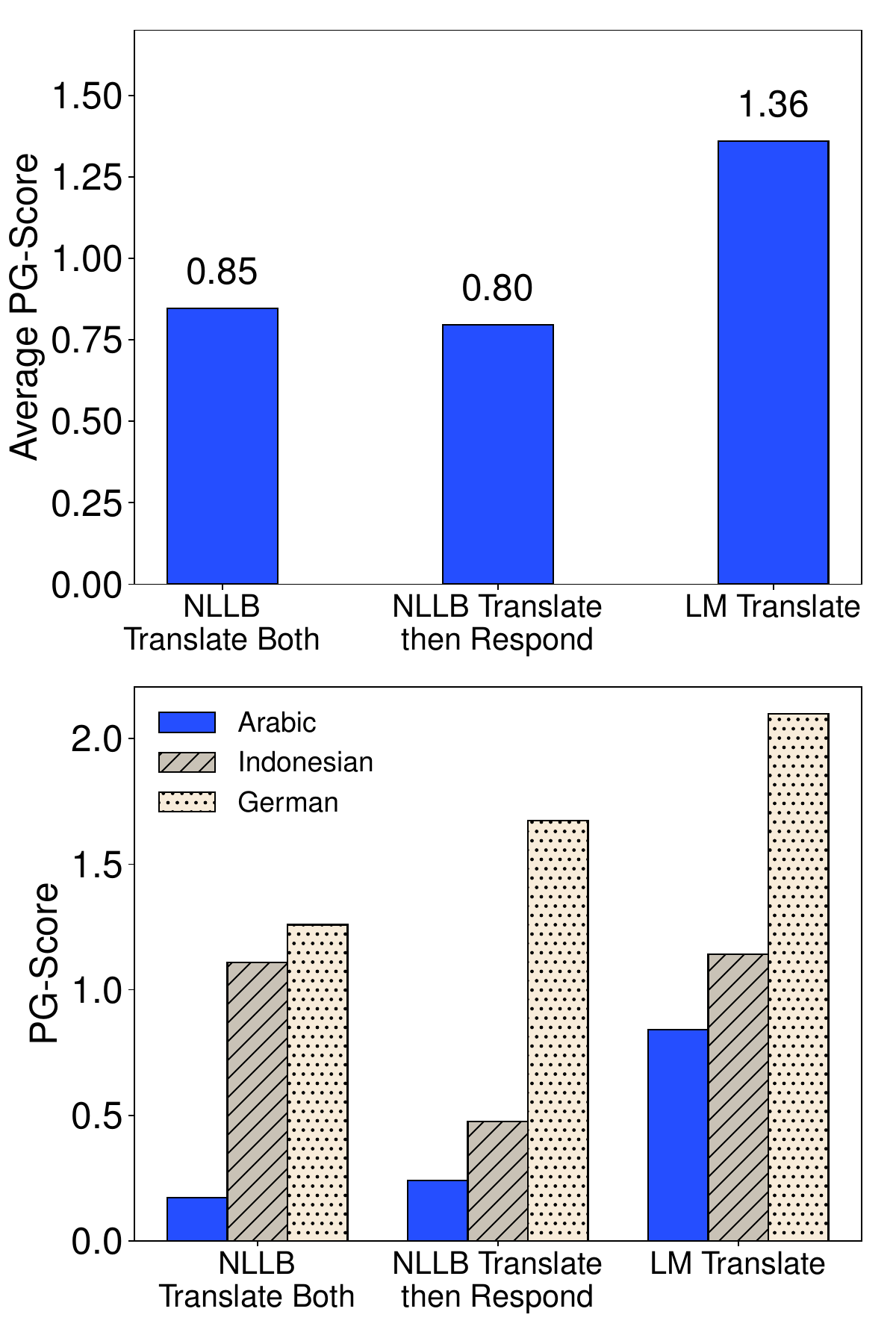}
    \caption{
        \textbf{Effect of translation method on \shortmethodname{}.}
        We compare three methods: LM translates prompt \texttt{EN-to-XX} and responds (\textit{LM-Translate}), NLLB translates prompt \texttt{EN-to-XX} and LM responds (\textit{NLLB-Translate-then-Respond}), and NLLB translates both prompt and response (\textit{NLLB-Translate-Both}).
    }
    \label{figure:translation_method}
\end{figure}

\paragraph{Results}
\autoref{figure:translation_method} shows the \shortmethodname{} and average benchmark performance of the student model for each translation method across Arabic, German, and Indonesian.
We find that \textit{LM-Translate} outperforms both NLLB-based approaches, achieving an average \shortmethodname{} of 1.36 compared to 0.85 for \textit{NLLB-Translate-Both} and 0.80 for \textit{NLLB-Translate-then-Respond}.
This pattern holds across all three languages, with the largest gap observed for German (2.09 vs.\ 1.26/1.68).

Our findings suggest that \textbf{prompt naturalness, rather than response quality, is a bottleneck in translation-based pipelines}:
having an LM generate responses to NLLB-translated prompts provides no improvement over pure NLLB translation (0.80 vs 0.85), indicating that translated prompts fail to elicit the same quality of responses as LM-translated prompts.

\begin{figure}
    \centering
    \includegraphics[width=0.33\textwidth, trim={0cm 0cm 0cm 0cm}]{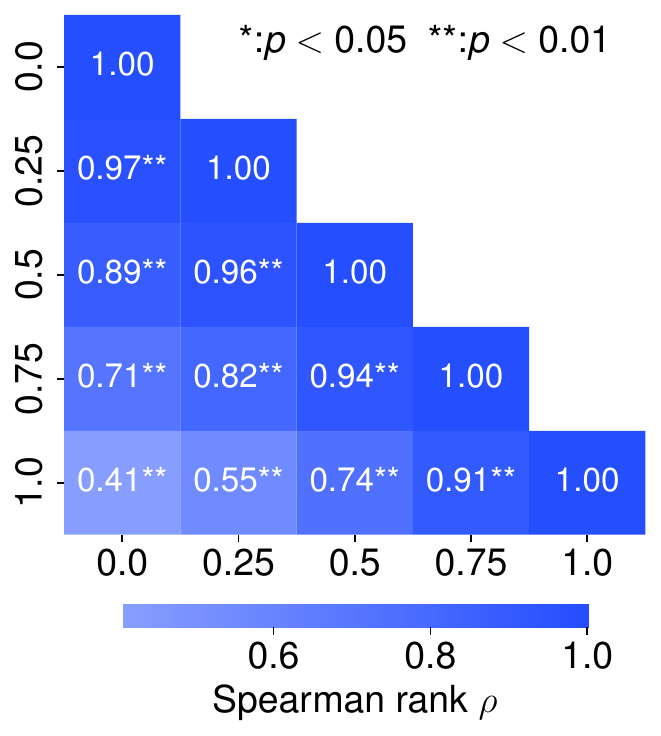}
    \caption{
        \textbf{Effect of weighting intrinsic and extrinsic metrics in \shortmethodname{}.}
        Model rankings remain relatively stable across neighboring weightings of intrinsic and extrinsic metrics.
    }
    \label{fig:pgscore_ablation}
\end{figure}

\subsection{Weighting of Intrinsic and Extrinsic Metrics in \shortmethodname{}}
\label{appendix:pgscore_ablation}

Our \shortmethodname{} formulation uses an assumption-free and equal weighting scheme between the intrinsic ($\mathcal{I}$) and extrinsic ($\mathcal{E}$) metrics.
In this section, we test whether these two metrics capture
(1) complementary aspects of teacher effectiveness and
(2) how model rankings differ if one metric is weighted more than the other.

\paragraph{Setup}
In order to test whether each metric captures complementary aspects of teacher effectiveness, we compute the Spearman rank correlation ($\rho$) between the intrinsic and extrinsic metrics across all teacher-language pairs (N=60, \nummodels{} models $\times$ \numlanguages{} languages).
In addition, in order to test the effect of weighting one metric against the other, we formulate a generalized version of \shortmethodname{}:

\vspace*{-1em}
\begin{equation}
    \begin{split}
        \text{\shortmethodname{}}_{T,\ell} & = \alpha \mathcal{I} + (1-\alpha) \mathcal{E} \\
                                           & \text{where } 0 \leq \alpha \leq 1
    \end{split}
\end{equation}

Note that the experiments in \S\ref{sec:experiments} and \S\ref{sec:analysis} assume $\alpha=0.5$.
We compute the \shortmethodname{} across $\alpha = \{0.00, 0.25, 0.50, 0.75, 1.00\}$ and then test the resulting model ranks' $\rho$ across all pairs of $\alpha$.
We perform this experiment on all teacher-language pairs where students are finetuned from the \basemodel{} base model (N=60, \nummodels{} models $\times$ \numlanguages{} languages).

\begin{table*}[t]
    \centering
    %\footnotesize
    \begin{tabular}{@{}lrrrr@{}}
        \toprule
                               & \multicolumn{4}{c}{\textbf{Generation Parameters}}                                                          \\
        \cmidrule(lr){2-5}
        \textbf{Model Name}    & \textbf{Temperature}                               & \textbf{Top-p} & \textbf{Top-k} & \textbf{Max Seq Len} \\
        \midrule
        GPT-4o mini            & 0.8                                                & 0.9            & --             & 16,384               \\
        Llama 3.1 70B Instruct & 0.6                                                & 0.9            & --             & 131,072              \\
        Llama 3.1 8B Instruct  & 0.6                                                & 0.9            & --             & 131,072              \\
        Command A              & 0.3                                                & --             & --             & 128,000              \\
        Aya Expanse 32B        & 0.3                                                & --             & --             & 128,000              \\
        Gemma 3 27B Instruct   & 1.0                                                & 0.95           & 64             & 8,192                \\
        Gemma 3 12B Instruct   & 1.0                                                & 0.95           & 64             & 8,192                \\
        Gemma 3 4B Instruct    & 1.0                                                & 0.95           & 64             & 8,192                \\
        IBM Granite 4.0        & 0.0                                                & --             & --             & 4,096                \\
        IBM Granite Micro      & 0.0                                                & --             & --             & 4,096                \\
        \hdashline
        Default                & 0.8                                                & 0.9            & --             & --                   \\
        \bottomrule
    \end{tabular}
    \caption{
        \textbf{Inference settings for each teacher model.}
        Generation parameters are based on model provider recommendations from HuggingFace and/or official documentation.
        % For Llama 3.1, we use temperature=0.6 and top\_p=0.9 as recommended by Meta.
        % For Gemma 3 models, we use temperature=1.0, top\_k=64, and top\_p=0.95 as recommended by Google.
        % For Aya Expanse and Command A, we use temperature=0.3 as shown in Cohere's official examples.
        % For IBM Granite, we use temperature=0.0 as recommended by IBM for most inferencing tasks.
        The Default row indicates parameters used when model-specific recommendations are unavailable.
        The ``--'' symbol indicates the parameter was not specified in the official recommendations.
    }
    \label{table:inference_details}
\end{table*}

\paragraph{Results}
Intrinsic and extrinsic metrics show a moderate positive correlation (Spearman $\rho=0.41$, $p<0.01$), suggesting that \textbf{data quality metrics are predictive of student performance while capturing complementary information}.
This finding motivates our combined \shortmethodname{} computation.
In addition, teacher rankings are stable for nearby weighting schemes ($\rho \geq 0.90$ for adjacent $\alpha$ values) as shown in \autoref{fig:pgscore_ablation}.
Our finding suggests that \textbf{model rankings are robust to small changes in the weighting of intrinsic and extrinsic metrics}.
Our equal weighting ($\alpha=0.5$) balances both perspectives, correlating strongly with extrinsic-focused ($\rho=0.89$) and reasonably with intrinsic-focused ($\rho=0.74$) rankings.

\subsection{Effect of language resource levels on \shortmethodname{}}
\label{sec:language}

% The results in \autoref{table:sota_evals} suggest that \shortmethodname{} varies considerably across the six languages tested:
% Arabic has the lowest \shortmethodname{} across all models, while German consistently has the highest scores.
% In this section, we investigate which language properties influence teacher effectiveness.

\paragraph{Setup}
For each language, we consider the following properties drawn from prior work:
CommonCrawl (CC) percentage as a proxy for presence in pretraining data \citep[\% in CC,][]{raffel2023exploringlimitstransferlearning},
and linguistic resource availability \citep[score from 0--5, 5 as high-resource, obtained from the LDC Catalog and the ELRA Map,][]{joshi-etal-2020-state}.
We compute the Spearman rank correlation ($\rho$) between each property and \shortmethodname{} across all teacher-language pairs (N=60, \nummodels{} models $\times$ \numlanguages{} languages).

\begin{figure}[t]
    \centering
    \includegraphics[width=0.70\linewidth, trim={1cm 0.5cm 0cm 0cm}]{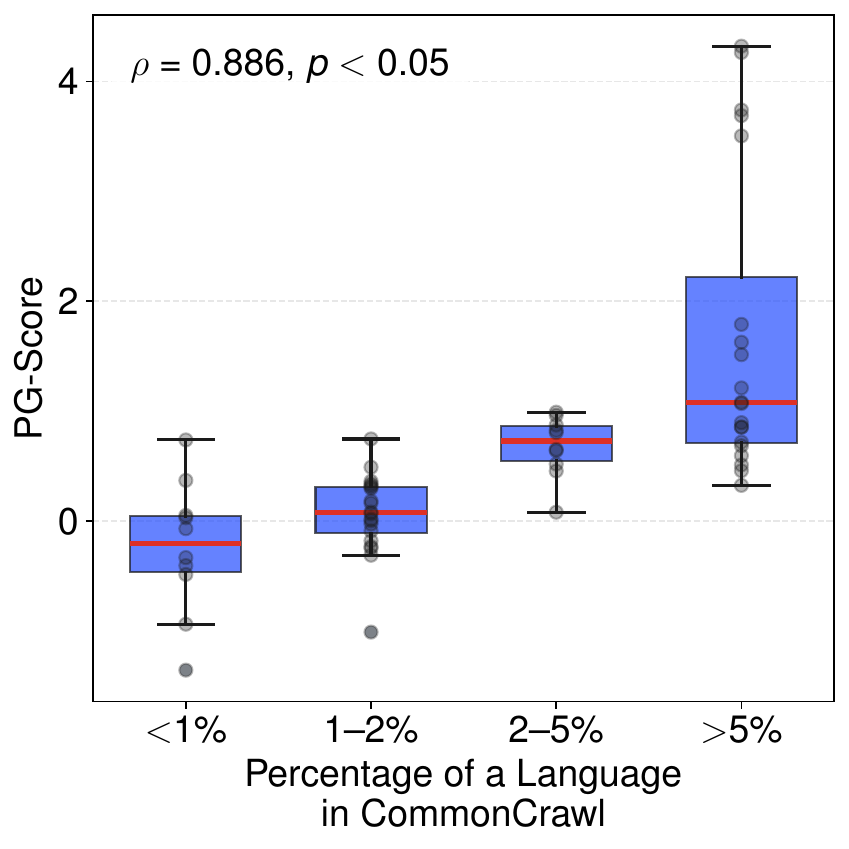}
    \caption{
        \textbf{Relationship between a language's percentage in CommonCrawl and \shortmethodname{}.}
        We observe a suggestive positive trend ($\rho=0.886$, $p<0.05$) between CommonCrawl representation and \shortmethodname{} across the six languages tested.
        % Across the six languages we tested, we observe that teacher models achieve higher \shortmethodname{} on languages with greater CommonCrawl representation and high-resource status.
    }
    \label{fig:language_correl}
\end{figure}

\paragraph{Results}
\autoref{fig:language_correl} shows the relationship between a language's percentage in CommonCrawl and \shortmethodname{}.
We observe a \textbf{suggestive positive trend between CommonCrawl representation and \shortmethodname{}} ($\rho=0.886$, $p<0.05$).
This finding suggests that languages with greater presence in pretraining data enable teacher models to generate higher-quality synthetic data that leads to better student performance.
This finding is unsurprising, but it provides empirical evidence of a structural gap that inhibits quality synthetic data generation for long-tail languages.
In contrast, we do not find a significant correlation between resource availability and \shortmethodname{} ($\rho=0.372$, $p=0.468$).
Our findings suggest that teacher model generation quality depends more heavily on pretraining exposure than linguistic resources.
Additionally, the data sources from \citet{joshi-etal-2020-state} do not reflect the current landscape: recent LMs are trained on either publicly available datasets from HuggingFace or in-house datasets.
While our work includes \numlanguages{} diverse languages, the sample size remains limited; we encourage future work to expand the number of languages to validate these findings.

\section{Disclosure on the Use of LLMs}

% good template for AI disclosure: https://arxiv.org/pdf/2608.23271
In this work, we used generative AI tools such as Claude Opus 4.6 for drafting parts of the paper, including title ideation and proofreading.
Additionally, we also used the same model to help with writing software code.
We have reviewed all AI-assisted work by proper code review and unit testing to check for correctness.
We also revised LLM-written text to improve readability and fit the author's writing style.
We take responsibility for the final content of this work, including text/claims/artifacts produced with the aid of generative AI.

\section{Multilingual Synthetic Data Recipe: Case Study on Tagalog}
\label{appendix:tagalog_case_study}

As an application of our findings and discussion in \S\ref{sec:discussion},
we present a case study on developing a multilingual synthetic data recipe on a held-out language: Tagalog.
It is a mid-resource language (Category 3 in \citet{joshi-etal-2020-state}'s taxonomy) and the standardized form of Filipino, the national language of the Philippines.

\subsection{Setup: Recipe Design and Evaluation}
\label{appendix:case_study_setup}

\paragraph{Data}
We collect Filipino seed data from various publicly available SFT datasets such as WildChat 4.8M and the Aya Collection.
In addition, we also include English data from the T\"ulu 3 SFT dataset for the \textit{Translate} method.
\autoref{table:tgl_seed_dataset} shows the statistics of the seed dataset used for Tagalog synthetic data generation.
Then, we implement the following data interventions based on our findings:

\begin{itemize}[complist]
    \item \textbf{Teacher Model:} we use Gemma 3 27B Instruct as the teacher model, as it was the best-performing model across most target languages we evaluated (\S\ref{sec:experiments}).
    \item \textbf{Data Generation Method:} we use the \textit{Translate} and \textit{Respond} methods, as they were the best-performing methods for mid-resource languages like Indonesian (\S\ref{sec:effect_generation_method}). In addition, we add a small sample of prompt-response pairs synthesized via the \textit{Generate} method.
    \item \textbf{Synthetic Data Scale:} we generate 10k synthetic examples using the selected teacher and data generation method, as we found that this scale is sufficient to achieve strong student performance (\appref{appendix:data_scale}).
          However, we also test on finetuning a model with 25k synthetic examples to see if more data improves performance.
    \item \textbf{Student Base Model:} we finetune using the Gemma 3 4B model, as we find that family-matched teacher-student pairs yield higher \shortmethodname{} (\S\ref{sec:generalization_base_models}).
\end{itemize}

\begin{table}[t]
    \centering
    % \footnotesize
    \renewcommand{\arraystretch}{0.9}
    \setlength{\tabcolsep}{1.2pt}
    % \resizebox{\textwidth}{!}{
    \begin{tabular}{@{}lr@{}}
        \toprule
        \textbf{Source} & \textbf{Num. Instances} \\
        \midrule
        %\begin{noindent}
        TaCo Alpaca     & 10,000                  \\
        Aya Collection  & 1,241                   \\
        WildChat 4.8M   & 997                     \\
        WildChat 1M     & 250                     \\
        \bottomrule
    \end{tabular}
    % }
    \caption{
        \textbf{Tagalog seed dataset statistics.}
        In order to bootstrap the synthetic data generation recipe for Tagalog, we curate a seed dataset containing a mix of Tagalog and English prompts from various sources.
        The majority of the seed dataset is from the TaCo paper \citep{upadhayay2024tacoenhancingcrosslingualtransfer}.
    }
    \label{table:tgl_seed_dataset}
\end{table}

For the purposes of this work, we designate the model finetuned on Gemma 3 4B using our synthetic recipe as 10K-Polyglot-TL, where ``10K'' indicates the number of SFT instances used during finetuning.

\paragraph{Evaluation}
We evaluate on \textsc{FilBench} \citep{miranda-etal-2025-filbench}, a benchmark for LMs that includes Filipino-centric multiple-choice and generative tasks.
It measures an LM's performance across four categories: classical NLP, cultural knowledge, reading comprehension, and generation, alongside an aggregated \textsc{FilBench} score.

We also compare against two data mix baselines:
\begin{enumerate}[complist]
    \item \textbf{10K-Public}: we sample 10k Tagalog prompt-response pairs from the seed dataset.
          This baseline aims to simulate a non-synthetic data approach to training multilingual LMs.
    \item \textbf{10K-GPT-4oM}: we synthesize 10k instances using an off-the-shelf teacher
          model (GPT-4o mini). This baseline simulates a typical data generation approach of choosing
          a teacher in an ad hoc manner due to its perceived strength (size or
          benchmark performance) or ease of use.
\end{enumerate}

For all methods, we finetune a Gemma 3 4B base model using the same training settings indicated in \appref{appendix:finetuning_details}.

\subsection{Results: Leaderboard Scores and Ablations}
\label{appendix:case_study_leaderboard}

\definecolor{cambridgelightblue}{HTML}{DCE3FF}
\definecolor{slate}{HTML}{ECEEF1}

\begin{table}[t]
    \centering
    % \resizebox{\linewidth}{!}{
    \begin{tabular}{@{}l@{\hspace{0.8em}}r@{}}
        \toprule
        \textbf{Model}                                            & \textbf{\textsc{FilBench} Score} \\
        \midrule
        GPT-4o (2024-08-06)                                       & 74.27                            \\
        Gemma 3 27B Inst.                                         & 55.17                            \\
        Gemma 3 12B Inst.                                         & 54.04                            \\
        \rowcolor{cambridgelightblue} \textbf{25K-Polyglot-TL 4B} & 49.73                            \\
        \rowcolor{cambridgelightblue} \textbf{10K-Polyglot-TL 4B} & 49.52                            \\
        Qwen 3 4B                                                 & 48.42                            \\
        \rowcolor{slate} 10K-GPT-4oM                              & 47.67                            \\
        Llama 3.1 8B Inst.                                        & 47.38                            \\
        Ministral 8B Inst.                                        & 47.33                            \\
        \rowcolor{slate} 10K-Public                               & 47.24                            \\
        Pangea 7B                                                 & 43.98                            \\
        SeaLLMs 3 1.5B                                            & 43.20                            \\
        \bottomrule
    \end{tabular}
    % }
    \caption{
        \textbf{Model performance on a held-out language (Tagalog) as evaluated on \textsc{FilBench}} \citep{miranda-etal-2025-filbench}.
        We compare \colorbox{cambridgelightblue}{our optimal synthetic recipe} against \colorbox{slate}{baseline approaches}
        and other models in the same parameter range.
    }
    \label{table:filbench_leaderboard}
\end{table}

\autoref{table:filbench_leaderboard} shows the \textsc{FilBench} score of our optimal synthetic recipe compared to other models in the same parameter range.
We find that 10K-Polyglot-TL outperforms both 10K-GPT-4oM ($+1.85$\,pp) and 10K-Public ($+2.28$\,pp).
These results suggest that (1) synthetic data generation is a viable approach for building models for less-resourced languages, and (2) selecting teacher models based on \shortmethodname{} is effective, as larger models do not always produce better training data (\S\ref{sec:experiments}).

In addition, comparing 10K-Polyglot-TL to other models in the \textsc{FilBench} leaderboard\footnote{Official \textsc{FilBench} leaderboard: \url{https://hf.co/spaces/filbench/filbench-leaderboard}} shows that our model is competitive against Qwen 3 4B and Llama 3.1 8B Instruct.
We highlight that \textbf{our 4B models are competitive against other models with larger parameter sizes}, suggesting that a multilingual synthetic data recipe based on our \shortmethodname{} findings is data-efficient.
We also find that increasing the number of SFT instances (10k to 25k) led to a performance increase of $0.21$\,pp.
While we previously found that gains diminish beyond 10k instances (see \appref{appendix:data_scale}), the continued gains from scaling to 25k instances on \textsc{FilBench} suggest that saturation points may depend on task diversity.
\textsc{FilBench} covers a broader range of NLP tasks (e.g., named-entity recognition) compared to our experimental benchmarks in \S\ref{sec:experiments} and \appref{appendix:additional_experiments}, indicating that \textbf{practitioners working with diverse task distributions may benefit from exploring larger synthetic datasets beyond the 10k threshold}.

\subsection{Analysis: Ablation Experiments}
In order to measure the contribution of our findings and recommendations in \S\ref{sec:discussion}, we perform the following ablation experiments as shown in \autoref{fig:tgl_ablation_filbench_scores}.
Note that the interventions described below are additive.

% reminder to LJ:
% - to download the results, please run this:
% - scripts/tagalog_ablations/get_results_XX.sh: where XX is the experiment number.

\paragraph{Curation of publicly available data vs.\ synthetic data generation}
We compare student models trained on (1) publicly available Tagalog SFT data and (2) synthetic SFT instances generated by a GPT-4o mini teacher
(note that these are also the same baselines in \appref{appendix:case_study_leaderboard}).
We find that the performance of these two baselines is similar ($\Delta=0.43$\,pp), suggesting that there is no significant advantage to using a synthetic data pipeline if the teacher model is not optimal.
We also hypothesize that some publicly-accessible datasets in Tagalog were semi-synthetic (e.g., TaCo uses a synthetic pipeline akin to the \textit{Translate} method, but using chain-of-thought to improve the quality of translations), making it difficult to perform a fair comparison.

\paragraph{Using a teacher with a higher \shortmethodname{}}
We then swap the GPT-4o mini teacher with Aya Expanse 32B, a teacher with a higher \shortmethodname{} based on our main findings (0.706 vs.\ 0.461, cf.\ \S\ref{sec:experiments}, \autoref{table:sota_evals}).
We observe a slight performance improvement in this intervention, suggesting that the \shortmethodname{} metric is generalizable across an unseen language.

\paragraph{Matching teacher and student model families}
One of our key findings and recommendations is to match the model families of the teacher and the student (\S\ref{sec:generalization_base_models}).
We use a Gemma 3 27B Instruct teacher model to match the family of the Gemma 3 4B base model.
This intervention yields a substantial performance improvement, consistent with our finding that family matching is a useful default for teacher selection (\S\ref{sec:generalization_base_models}).
We attribute this to shared tokenization schemes and architectural similarities that facilitate better knowledge transfer from teacher to student.

\paragraph{Increase data scale}
We increase the number of synthetic instances from 10k to 25k to assess whether additional data continues to improve performance.
We observe a modest gain of $0.21$\,pp, which is smaller than the improvements from teacher model selection and model family matching.
This finding aligns with our earlier observation that gains diminish beyond 10k examples (\appref{appendix:data_scale}), though the continued improvement on \textsc{FilBench}'s diverse task distribution suggests that saturation points may be task-dependent.

\paragraph{Increase model scale}
Finally, we explore whether scaling the student model from 4B to 12B and 27B parameters provides additional performance gains.
We find that the larger student model achieves higher performance, demonstrating that our synthetic data recipe benefits from increased model capacity.
This result is consistent with our generalization experiments (\appref{appendix:model_scale}), where we showed that \shortmethodname{} generalizes across different model sizes while maintaining the relative ranking of teacher models.
However, we note that the performance of our best models is still behind Gemma 3 27B Instruct and Gemma 3 12B Instruct (\autoref{table:filbench_leaderboard}).
Given that observation, we still argue that our synthetic pipeline, which uses 25k instances trained only via SFT, can be considered data- and resource-efficient compared to the post-training interventions done in Gemma 3, which involved instruction-tuning and reinforcement learning objectives \citep{team2025gemma}.

\subsection{Analysis: Instance-Level Qualitative Inspection of Generated Data}

\paragraph{Setup}
We also inspect the quality of generated Tagalog data across teacher models with different \shortmethodname{}s.
To do so, we sample 50 instances from a dataset generated by the following models (in increasing \shortmethodname{}):
GPT-4o mini, % (\shortmethodname{}\textsubscript{tl}=0.480), 
Aya Expanse, % (\shortmethodname{}\textsubscript{tl}=0.822), 
and Gemma 3 27B.% (\shortmethodname{}\textsubscript{tl}=1.049).
We evaluate these instances in terms of the following criteria:

\begin{itemize}[complist]
    \item \textit{Fluency}: whether the generated text reads naturally in Tagalog, with grammatical correctness and idiomatic word choice rather than literal translations or code-switched constructions that a native speaker would find awkward.
    \item \textit{Cultural accuracy}: whether the content reflects appropriate Filipino cultural context, including correct references to local entities, customs, and norms, as opposed to generic or Western-centric framings that have been superficially translated into Tagalog.
\end{itemize}

The scoring is from 1--5, using the rubric:

\begin{enumerate}[complist]
    \item \textbf{Unacceptable}: ungrammatical, code-switched, or reads as a literal translation (\textit{fluency}); content is absent, generic, or Western-centric (\textit{cultural accuracy}).
    \item \textbf{Poor}: frequent grammatical or idiomatic errors (\textit{fluency}); cultural references are mostly missing or incorrect (\textit{cultural accuracy}).
    \item \textbf{Adequate}: comprehensible but noticeably awkward (\textit{fluency}); some appropriate grounding mixed with generic or Western framings (\textit{cultural accuracy}).
    \item \textbf{Good}: mostly natural with only minor lapses (\textit{fluency}); cultural context is largely accurate with small inaccuracies (\textit{cultural accuracy}).
    \item \textbf{Excellent}: fully fluent and idiomatic, indistinguishable from native writing (\textit{fluency}); consistently and accurately grounded in Filipino cultural context (\textit{cultural accuracy}).
\end{enumerate}

These annotations were performed by a native speaker who is one of the authors of this work. 
To reduce bias, the annotator only saw a prompt-response pair and its generation method without knowledge of which teacher model it was distilled from.
Finally, we also include some observations by the annotator during the annotation process.

\paragraph{Results}
\autoref{table:qualitative_evaluation} shows the annotation results for the 50 instances sampled from each teacher model.
We find that for Tagalog, 
teacher models with a higher \shortmethodname{} also produce higher-quality Tagalog data as judged by a native speaker.
Gemma 3 27B, which has the highest \shortmethodname{}\textsubscript{tl} (1.049), 
achieves the strongest ratings on both criteria.
This pattern is most consistent on cultural accuracy, 
which increases monotonically with \shortmethodname{}, suggesting that \shortmethodname{} captures aspects of cultural grounding that are perceptible to native speakers.

However, we observe that fluency does not follow the same trend: Aya Expanse receives a lower fluency score than GPT-4o mini (2.480 vs.\ 2.800) despite a higher \shortmethodname{}\textsubscript{tl}.
We also note the substantial gap between Gemma 3 27B and the other two teachers on both dimensions 
($\Delta=1.3$--$1.6$ on fluency, $\Delta=0.9$--$1.5$ on cultural accuracy), 
consistent with our earlier finding that matching the teacher model family with the student is a useful default for teacher selection (\S\ref{sec:generalization_base_models}).

\begin{table}[t]
    \centering
    \begin{tabular}{lcc}
        \toprule
        \textbf{Teacher Model} & \textbf{Fluency} & \textbf{Cul. Acc.} \\
        \midrule
        GPT-4o mini     & 2.800 & 2.740  \\
        Aya Expanse     & 2.480 & 3.320 \\
        Gemma 3 27B     & 4.100 & 4.220 \\
        \bottomrule
    \end{tabular}
    \caption{
        \textbf{Native-speaker annotation of teacher-generated outputs.}
        We show the average across 50 instances of the native-speaker ratings on Tagalog synthetic data.
    }
    \label{table:qualitative_evaluation}
\end{table}

\section{Inference Details}
\label{appendix:prompt_templates}

\paragraph{Prompt templates}
\autoref{fig:generate_prompt_template} to \autoref{fig:respond_prompt_template} show the prompt templates used for each data generation method.
In addition, \autoref{fig:llm_as_judge_prompt} shows the prompt template used for the LLM-as-a-judge method to evaluate text quality.

\paragraph{Inference settings}
We use vLLM \citep{kwon2023efficient} and Curator \citep{marten2025curator} for inference.
For each teacher model, we check whether the model provider recommended settings for usage.
If not, then we set a default configuration (temperature=0.8, top\_p=0.9).
\autoref{table:inference_details} summarizes the inference settings we used for each teacher model.

\begin{figure*}[t]
   % \footnotesize
   \small
    \centering
    \begin{minipage}{0.95\textwidth}
        \promptbox[Generate: sample $k$ prompt-response pairs from $\seeddata$ and use it as in-context examples]{
        As a multilingual data generator, your task is to generate a new example (\texttt{`prompt`} and \texttt{`response`}) for a dataset demonstrating how AI agents can fulfill general instructions for \texttt{\{lang\_name\}}.\\\\
        To do this, you will want to generate two pieces of information:\\
        1) A "prompt" specifying a task to be completed or a question to be answered (what, where, when, how, who, why). The task should be very challenging yet solvable.\\
        2) A "response" representing a valid completion of that task in natural language. If the "response" does not satisfy the "prompt", then you have failed at your job. Do not provide unnecessary details, beyond what is explicitly needed to satisfy the instruction you generated.\\
        \\
        Hard constraint: The generated task MUST belong to exactly one of the following categories (pick one at random and do NOT mention the category).\\
        1. Logical reasoning / error analysis\\
        2. Math or quantitative reasoning with explanation\\
        3. Classification or labeling\\
        4. Dialogue or role-play\\
        5. Translation or paraphrasing with constraints\\
        6. Procedural instructions (step-by-step)\\
        7. Grammar correction or linguistic analysis\\
        8. Short-form creative output ($\leq$50 words)\\
        9. Knowledge recall with verification or correction\\
        10. Cultural or pragmatic judgment\\
        \\
        Add diversity to your generations by varying the types of tasks you create, the styles and tones of the responses, and the complexity of the language used. This will help ensure a rich and varied dataset.
        For example, you might create tasks that involve answering knowledge-based questions, answering math questions, providing explanations, generating creative content, or performing translations.
        \\
        \\
        Please provide a JSON dictionary response that includes the new \texttt{`prompt`} and its corresponding \texttt{`response`}. Use the \texttt{`prompt`} and \texttt{`response`} keys in the dictionary.\\
        Do not generate any other text in your response (for example, do not start your message with any greetings, and never ask for clarification or apologize for struggling with the task).\\
        Try your best to ensure that the input and response you generate are distinct from the provided examples while maintaining a diverse, detailed, precise, comprehensive, and high-quality response.\\
        It is important to generate responses that are contextually relevant and culturally appropriate for \texttt{\{lang\_name\}}.\\
        \\
        Here are some examples to guide your generation. The best way to use these examples is to identify the patterns and structures they follow, rather than copying them directly:\\
        \\
        \texttt{\{\% for example in examples[:k] \%\}}\\
        \texttt{Prompt: \{\{example[``prompt'']\}\}}\\
        \texttt{Response: \{\{example[``response'']\}\}}\\
        \texttt{\{\% endfor \%\}}\\
        \\
        New Example:
        }
    \end{minipage}
    \caption{Prompt template for the \textit{Generate} data generation method.}
    \label{fig:generate_prompt_template}
\end{figure*}

\begin{figure*}[t]
    %\small
    %\footnotesize
    \centering
    \begin{minipage}{0.95\textwidth}
        \promptbox[Translate: forward-translate English prompts from $\seeddata$ and use teacher $\teacher$ to generate the response $y_i$]{
        As a multilingual data generator, your task is to translate the given prompt from English into \texttt{\{lang\_name\}} and generate the appropriate response in the same language.\\
        Important: you must return both the translated prompt (into \texttt{\{lang\_name\}}) and the response. Ensure that both the translated prompt and the response are coherent, culturally appropriate, and demonstrate a deep understanding of the language nuances.\\
        \\
        Do not generate any other text in your response (for example, do not start your message with any greetings, and never ask for clarification or apologize for struggling with the task).\\
        Do not return the original English prompt. Remember, you must translate the prompt first and return it.\\
        Here is the prompt you need to translate and respond to:\\
        \\
        \texttt{\{prompt\}}
        }
    \end{minipage}
    \caption{Prompt template for the \textit{Translate} data generation method.}
    \label{fig:translate_prompt_template}
\end{figure*}

\begin{figure*}[t]
    %\small
    %\footnotesize
    \centering
    \begin{minipage}{0.95\textwidth}
        \promptbox[Respond: take prompts from $\seeddata$ and use teacher $\teacher$ to generate the response $y_i$]{
        As a multilingual data generator, you will be presented a user request or instruction in the \texttt{\{lang\_name\}} language. Your task is to generate an appropriate response for the given request.
        Ensure that your response is coherent, culturally appropriate, and demonstrates a deep understanding of the language nuances
        Do not generate any other text in your response (for example, do not start your message with any greetings, and never ask for clarification or apologize for struggling with the task).
        Here is the prompt you need to respond to:\\
        \\
        \texttt{\{prompt\}}
        }
    \end{minipage}
    \caption{Prompt template for the \textit{Respond} data generation method.}
    \label{fig:respond_prompt_template}
\end{figure*}

\begin{figure*}[t]
    %\footnotesize
    \centering
    \begin{minipage}{0.95\textwidth}
        \promptbox[LLM-as-a-judge: evaluating text quality using the multilingual rubric language model]{
            \textbf{Task Description:}\\
            An instruction (might include an Input inside it) in \texttt{\{language\}}, a response to evaluate, and a score rubric representing a evaluation criteria are given.\\
            1. Write a detailed feedback that assess the quality of the response strictly based on the given score rubric, not evaluating in general.\\
            2. After writing a feedback, write a score that is an integer between 1 and 5. You should refer to the score rubric.\\
            3. The output should contain the score and feedback only.\\
            4. Please do not generate any other opening, closing, and explanations.\\

            \textbf{The instruction to evaluate:}\\
            \texttt{\{\{instruction\}\}}
            \\
            \\
            \textbf{Response to evaluate:}\\
            \texttt{\{\{response\}\}}
            \\
            \\
            \textbf{Score Rubrics:}

            [Is the model proficient in language \texttt{\{lang\_name\}}, including its cultural nuance and grammatical usage, and responds in a helpful and harmless manner according to the instruction?]

            Score 1: The response contains severe grammatical errors, lacks cultural appropriateness, or is unhelpful/harmful. The language proficiency is very poor.\\
            Score 2: The response has noticeable grammatical errors and limited cultural awareness. It partially addresses the instruction but with significant gaps in language proficiency or helpfulness.\\
            Score 3: The response demonstrates adequate language proficiency with some minor grammatical errors. It shows reasonable cultural awareness and addresses the instruction in a helpful manner, though improvements are possible.\\
            Score 4: The response exhibits strong language proficiency with minimal grammatical errors and good cultural nuance. It addresses the instruction in a helpful and harmless way with only minor room for improvement.\\
            Score 5: The response demonstrates excellent language proficiency with proper grammar, appropriate cultural nuance, and idiomatic usage. It fully addresses the instruction in a helpful and harmless manner.\\
            \\
            \\
            \textbf{Feedback:}\\
        }
    \end{minipage}
    \caption{
        We evaluate text quality of synthesized texts using a multilingual rubric model called M-Prometheus \citep{pombal2025mprometheus}.
        We choose M-Prometheus due to its strong performance on multilingual and human-aligned benchmarks.
    }
    \label{fig:llm_as_judge_prompt}
\end{figure*}

% \clearpage
% \input{tables/absolute_benchmark_scores.tex}

\end{document}